\documentclass[10pt,twocolumn,letterpaper]{article}

\usepackage{cvpr}              % To produce the CAMERA-READY version
\definecolor{cvprblue}{rgb}{0.21,0.49,0.74}
\usepackage[pagebackref,breaklinks,colorlinks,allcolors=cvprblue]{hyperref}

\def\paperID{13705} % *** Enter the Paper ID here
\def\confName{CVPR}
\def\confYear{2026}

\usepackage{pifont}
\usepackage{multirow} % Required for multirows
\usepackage{amssymb}
\usepackage[ruled,vlined]{algorithm2e}
\usepackage[table,xcdraw]{xcolor}
\def\eg{{\it{e.g.}}}

\def\ie{{\it{i.e.}}}

\newcommand{\cmark}{\ding{51}}%
\newcommand{\xmark}{\ding{55}}%
\usepackage{listings}
\lstdefinestyle{promptstyle}{
  backgroundcolor=\color{gray!10},
  basicstyle=\ttfamily\small,
  breaklines=true,
  frame=single,
  rulecolor=\color{gray!40},
  columns=fullflexible,
}

\title{Explore with Long-term Memory: A Benchmark and Multimodal LLM-based Reinforcement Learning Framework for Embodied Exploration
}

\author{
Sen Wang\textsuperscript{1*}\quad
Bangwei Liu\textsuperscript{1}\quad
Zhenkun Gao\textsuperscript{1}\quad
Lizhuang Ma\textsuperscript{1}\quad
Xuhong Wang\textsuperscript{2}\quad
Yuan Xie\textsuperscript{1}\quad
Xin Tan\textsuperscript{1,2$\dagger$}\\[0.5em]
\textsuperscript{1}East China Normal University \quad
\textsuperscript{2}Shanghai AI Laboratory
}

\begin{document}
\twocolumn[{%
\renewcommand\twocolumn[1][]{#1}%
\maketitle
\vspace{-3em}
\begin{center}
    \centering
    \captionsetup{type=figure}
    \includegraphics[width=0.92\textwidth]{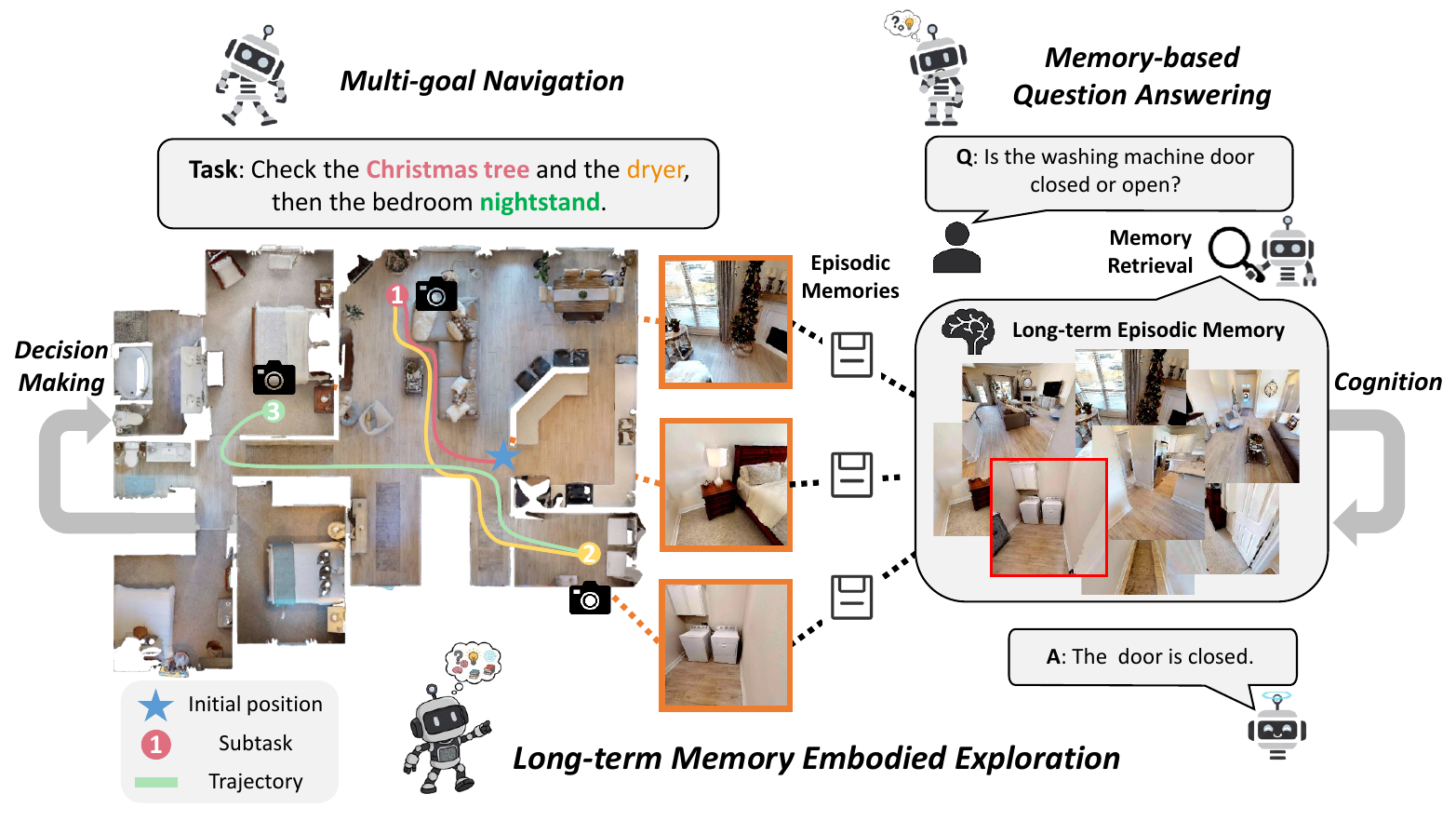}
    \captionof{figure}{We propose Long-term Memory Embodied Exploration, which aims to collect episodic memories during Multi-goal Navigation and introduces Memory-based Question Answering to unify and evaluate the model’s cognitive and decision-making abilities.}
\label{fig:intro_1}
\end{center}%
}]
\begingroup
\renewcommand\thefootnote{$\dagger$}\footnotetext{
Corresponding author.
}
\renewcommand\thefootnote{*}\footnotetext{
This work was done by Sen Wang during an internship at Shanghai AI Laboratory.
}
\endgroup

\begin{abstract}
An ideal embodied agent should possess lifelong learning capabilities to handle long-horizon and complex tasks, enabling continuous operation in general environments. This not only requires the agent to accurately accomplish given tasks but also to leverage long-term episodic memory to optimize decision-making. However, existing mainstream one-shot embodied tasks primarily focus on task completion results, neglecting the crucial process of exploration and memory utilization. To address this, we propose Long-term Memory Embodied Exploration (LMEE), which aims to unify the agent’s exploratory cognition and decision-making behaviors to promote lifelong learning. We further construct a corresponding dataset and benchmark, LMEE-Bench, incorporating multi-goal navigation and memory-based question answering to comprehensively evaluate both the process and outcome of embodied exploration. To enhance the agent’s memory recall and proactive exploration capabilities, we propose MemoryExplorer, a novel method that fine-tunes a multimodal large language model through reinforcement learning to encourage active memory querying. By incorporating a multi-task reward function that includes action prediction, frontier selection, and question answering, our model achieves proactive exploration. Extensive experiments against state-of-the-art embodied exploration models demonstrate that our approach achieves significant advantages in long-horizon embodied tasks. Our dataset and code will be released at \url{https://wangsen99.github.io/papers/lmee/}.

\end{abstract}    
\section{Introduction}
\label{sec:intro}

\begin{table*}[!t]
  \centering
  \caption{Comparison to popular embodied exploration benchmarks. }
  \vspace{-0.6em}
  \renewcommand{\arraystretch}{1.2}
   \resizebox{0.96\linewidth}{!}{
    \begin{tabular}{l|c|c|c|c|c|c|c|c|c}
    \toprule
    \multicolumn{1}{l|}{\multirow{2}[4]{*}{Dataset}} & \multicolumn{2}{c|}{Train} & \multicolumn{2}{c|}{Test} & \multirow{2}[4]{*}{Multi-goal} & \multirow{2}[4]{*}{Memory} & \multirow{2}[4]{*}{Question Num} & \multicolumn{2}{c}{Question Type} \\
\cmidrule{2-5}\cmidrule{9-10}          & Categories & Goals & Categories & Goals &       &       &       & Choices & Open Vocab \\
    \midrule
    ObjectNav-MP3D~\cite{chang2017matterport3d} & 21    & 7509  &-     &-  & \xmark &\xmark       &-   &-  &-  \\
    ObjectNav-HM3D~\cite{habitatchallenge2023} & 6     & 5216  &-   &-   &\xmark   & \xmark  &-  &-       &-  \\
    InstanceImageNav-HM3D~\cite{krantz2022instance} & 6     & 2516  &-       &-       &\xmark       &\xmark       &-       &-       &-  \\
    HM3D-OVON~\cite{yokoyama2024hm3d}  & 280   & 10987 & 49    & 1278  &\xmark       &\xmark       &-       &-       &-  \\
    GOAT-Bench~\cite{khanna2024goat} & 193   & 13025 & 36    & 1282  & \cmark      &\xmark       &-       &-       &-  \\
    MP3D-EQA~\cite{wijmans2019embodied} &-       &-       & -      &-       &\xmark       &\xmark       & 1136   & \xmark      & \xmark \\
    HM-EQA~\cite{ren2024explore} &-       &-       & -      &-       &\xmark       &\xmark       & 500   & \cmark      & \xmark \\
    OpenEQA~\cite{majumdar2024openeqa} &-       &-       &-       &-       &\xmark       &\xmark       & 1600+ &\xmark       & \cmark \\
    EXPRESS-Bench~\cite{jiang2025beyond} &-       &-       &-       & -      &\xmark       &\xmark      & 2044  &\xmark       & \cmark \\
    MemoryEQA~\cite{zhai2025memory} &-       &-       &-       &-       &\xmark       &  \cmark     & 1587  & \cmark      &\xmark  \\
    \rowcolor{blue!10}
    LMEE (Ours) & 208   & 8880  & 38    & 828  & \cmark      &  \cmark & 9286 (406) & \cmark      & \cmark  \\
    \bottomrule
    \end{tabular}%
    }
\vspace{-1em}
  \label{tab:intro_1}%
\end{table*}%

A key objective in embodied intelligence is to empower agents with lifelong learning capabilities, enabling them to perform complex tasks and operate continuously in dynamic and unfamiliar environments. For example, as illustrated in \cref{fig:intro_1}, consider the instruction ``\textit{Check the Christmas tree and the dryer, then the bedroom nightstand.}" After completing this sequence of tasks, the agent should have developed a comprehensive understanding of the explored environment. Later, when asked a follow-up question such as ``\textit{Is the washing machine door closed or open?}", the agent should be able to quickly retrieve its stored memory and respond, ``\textit{The door is closed.}" This capability represents more than simple task completion, as it demonstrates the agent’s ability to construct dynamic and context-aware memory representations that support efficient recall and reasoning for future interactions. Such an ability is essential for developing embodied agents that can continuously learn, adapt, and evolve in complex real-world environments~\cite{khanna2024goat}.

Embodied exploration aims to enable agents to proactively explore unknown environments. As shown in \cref{tab:intro_1}, current research paradigms mainly focus on tasks such as goal navigation~\cite{habitatchallenge2023,chang2017matterport3d,krantz2022instance,yokoyama2024hm3d} and embodied question answering~\cite{ren2024explore,majumdar2024openeqa,jiang2025beyond,zhai2025memory}. However, these one-shot tasks tend to emphasize outcomes while neglecting the exploration process itself. Although multi-goal navigation~\cite{khanna2024goat} focuses on long-horizon embodied tasks, it still overlooks how the exploration process contributes to the agent’s scene understanding and decision-making. Achieving a unified integration of cognition and decision-making is therefore crucial for developing general embodied intelligence.

Multimodal Large Language Models (MLLMs)~\cite{meta2024llama, li2024llava, team2025qwen3, jin2025efficient} have shown remarkable potential in embodied exploration, particularly for complex scenes~\cite{xie2023boosting, xie2024csfwinformer, zhang2025embodied, wang2026mmofusion, wang2025enhancement}. However, existing approaches still struggle to make effective use of memory. 
Many methods treat memory passively, limiting the agent's autonomy and reasoning capabilities. For example, imitation learning-based approaches~\cite{zhu2025move, zhang2025embodied} train agents to replicate expert trajectories. This passive learning paradigm restricts generalization to unseen scenarios and, crucially, prevents the agent from developing its own proactive exploration strategies. 
Other vision-language exploration methods~\cite{yang20253d, zhai2025memory} that depend on memory snapshots use filtering strategies to mitigate the constraints of limited context windows but fail to harness the active querying capability inherent in MLLMs. 
Likewise, models with long-term spatiotemporal memory~\cite{hu20253dllm} mainly perform post-exploration reasoning, missing the opportunity to use memory to proactively guide exploration.

To address these challenges, firstly, we introduce Long-term Memory Embodied Exploration (LMEE), which emphasizes not only the outcome (goal) of the embodied task but also the exploration process (memory). As illustrated in \cref{fig:intro_1}, LMEE consists of two key components: Multi-goal Navigation and Memory-based Question Answering. During navigation, the agent dynamically constructs on-the-fly memories. LMEE further provides a large collection of exploration-related questions that the agent must answer based on its memories.
To evaluate LMEE, we construct a comprehensive dataset and establish a benchmark, LMEE-Bench, which assesses the agent from two perspectives: (1) success rate and efficiency in multi-goal navigation, and (2) accuracy in memory-based question answering. This assesses the agent’s ability to use episodic memory. The dataset encompasses 246 object categories, over 9,000 goals and questions, and 1,982 exploration trajectories.

Secondly, we propose MemoryExplorer, an MLLM-based framework for active exploration and memory retrieval trained via Reinforcement Fine-Tuning (RFT). The model proactively queries and invokes memory retrieval tools to access multimodal memory information based on current task instructions, multi-view observations, and goal-related questions. By analyzing its answers and assessing task progress, the agent gains an understanding of the current situation and plans subsequent actions accordingly.
Furthermore, we design a Multi-Task Reward function that integrates action and frontier prediction with question answering, effectively unifying scene understanding, memory utilization, and planning-based decision-making. This enables the model to tackle challenging tasks in complex environments. Extensive experiments demonstrate that our approach achieves superior performance in long-term memory embodied exploration, significantly enhancing the model’s capacity for autonomous exploration and active memory retrieval, which are key abilities for realizing lifelong learning in embodied agents.

In summary, our contributions are as follows:
\begin{itemize}

\item We introduce LMEE, a new paradigm for developing autonomous agents by unifying exploration with memory-based reasoning. We also present its corresponding benchmark, LMEE-Bench, to holistically evaluate agents using multi-goal navigation and question answering to assess the crucial abilities of memory utilization, cognitive understanding, and decision-making.
\item We propose MemoryExplorer, which uses reinforcement learning to enable active exploration and memory retrieval in unknown environments. By combining frontier prediction, action planning, and question answering, we design a multi-task reward function that optimizes the model’s policy for effective reasoning in complex scenes.
% \item We introduce Long-term Memory Embodied Exploration (LMEE) along with its benchmark LMEE-Bench, which dynamically construct a multimodal memory bank through multi-goal navigation and incorporate memory-based question answering. This unified framework allows agents to develop autonomous exploration behaviors while providing a systematic benchmark to evaluate their ability to integrate cognitive memory and decision-making in lifelong learning scenarios.
% \item We propose MemoryExplorer, which leverages reinforcement learning to enable the model to actively explore and retrieve information in unknown environments. By integrating frontier image and action prediction with question answering, we design a multi-task reward function that optimizes the model’s policy across multiple dimensions, facilitating effective handling of complex scene information.
\end{itemize}

\section{Related Work}
\label{sec:relate}

\subsection{Embodied Navigation and Question Answering}
Goal-driven Navigation and Vision-Language Question Answering are mainstream tasks in embodied intelligence. Early navigation tasks mostly targeted single targets~\cite{wijmans2019dd, chaplot2020object, hahn2021no, ramrakhya2022habitat}, and due to the limitations of current model generalization capabilities, navigation methods are based on modular designs and only employ modality-specific encoders. Similarly, the questions in embodied question answering tasks are relatively simple~\cite{das2018embodied, yu2019multi, wijmans2019embodied}. In recent years, MLLMs have shown impressive results in the field of embodied intelligence~\cite{gu2025doraemon, li2025compassnav, xi2024thp, gao2025styleshot, zhao2024learning}. Whether in open question answering~\cite{majumdar2024openeqa} or multimodal navigation~\cite{khanna2024goat,cao2025cognav,yin2025unigoal}, embodied tasks are moving towards a long-term, generalized paradigm~\cite{zhang2025embodied, cheng2025embodiedeval, qiao2025navbench}. However, current tasks focus on the outcome, such as whether the target object has been found~\cite{jiang2025beyond}, or whether the answer to the embodied question is accurate~\cite{ren2024explore}. This lack of attention to the process is not conducive to building lifelong learning agents~\cite{tan2025towards}. We introduce Long-term Memory Embodied Exploration, aiming to focus on the preservation of contextual memory during long-range task exploration, enabling the agent to think about problems based on experience like a human, ultimately achieving lifelong learning and self-evolution.

\subsection{Memory-based Agents}

Several works have explored memory mechanisms for language-based agents. For example, MemGPT~\cite{packer2023memgpt}, MemAgent~\cite{yu2025memagent}, and Mem-$\alpha$\xspace~\cite{wang2025mem} provide large language models with extended context via diverse memory systems, while ReasoningBank~\cite{ouyang2025reasoningbank} builds experience pools to support long-term tasks such as web browsing and software engineering. However, these text-based or discrete-state-based memory models are not directly applicable to embodied AI, where memory must capture spatio-temporal information from the physical world.

Memory is crucial for long-horizon embodied tasks, which are growing increasingly complex~\cite{yang2025embodiedbench}. Methods like MTU3D~\cite{zhu2025move}, 3D-Mem~\cite{yang20253d}, and 3DLLM-Mem~\cite{hu20253dllm} construct memory banks or snapshots to support navigation, object localization, and spatio-temporal reasoning. Yet, these approaches mostly rely on passive memory usage, \eg{}, imitating trajectory data~\cite{zhu2025move, hu20253dllm}. In contrast, we aim to enable active memory querying to enhance proactive exploration and task handling in embodied settings.

\subsection{Reinforcement Learning for Embodied AI}
The advantage of reinforcement learning lies in the agent’s ability to learn actively~\cite{singh2022ask4help}, and the strong generalization capability of LLMs further amplifies this benefit. Recent work leverages LLMs to automatically learn reward models from interaction data without manual annotation~\cite{sarukkai2024automated, chen2025scaling}. Reinforcement learning for training multimodal LLMs to solve embodied tasks has also gained increasing attention. \cite{szot2023large} trains perception and action jointly through real-time environmental interaction, enabling generalization across embodied tasks. \cite{ramrakhya2025grounding} allows agents to ask clarifying questions for ambiguous instructions, while \cite{tian2025seea} introduces reinforcement fine-tuning to support self-evolution of embodied agents. However, early approaches rely on overly simple task instructions~\cite{szot2023large}, and active questioning remains limited in scope~\cite{ramrakhya2025grounding}, restricting applicability to complex scenarios. Other methods emphasize only task completion~\cite{qi2025vln, gao2025octonav, tian2025seea} and overlook the task process, which hinders lifelong learning. In contrast, we emphasize the memory retrieval capability of embodied agents for complex long-horizon tasks, enabling better scene understanding and more efficient decision-making.
\section{Data Construction of LMEE}
\label{sec:dataset}

\begin{figure*}[!t]
\centering
\includegraphics[width=0.92\linewidth]{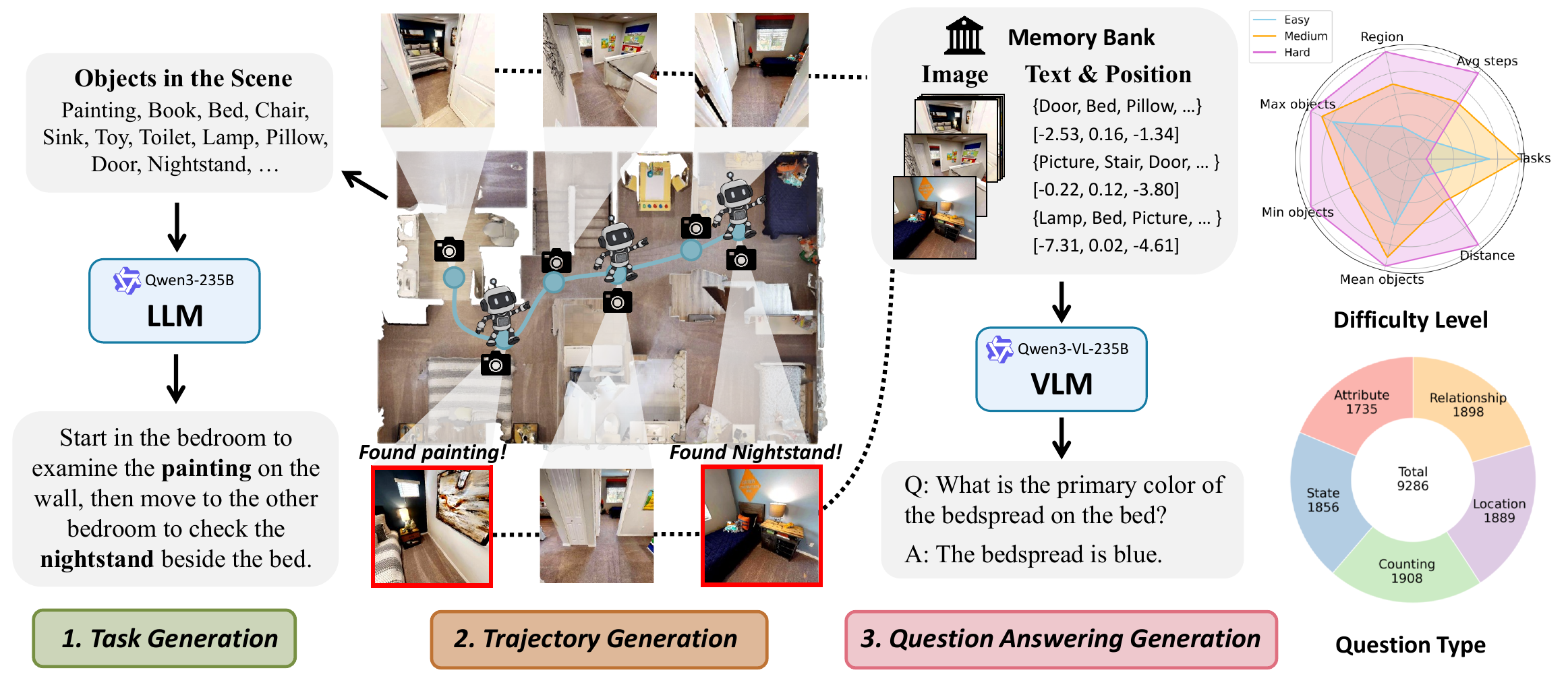}
\vspace{-0.6em}
\caption{The construction process of Long-term Memory Embodied Exploration and data statistics.}
\label{fig:data_1}
\vspace{-1.2em}
\end{figure*}

To enable agents to proactively explore and memorize in unknown environments, we construct a Long-term Memory Embodied Exploration dataset. Based on multi-goal navigation tasks, it dynamically builds a memory bank by collecting observations during the exploration process, thereby enabling memory-based embodied exploration training and evaluation. As shown in \cref{fig:data_1}, the data construction process consists of three parts: task instruction generation, exploration trajectory generation, and memory-based question answering generation.

\noindent \textbf{Task Instruction Generation.} We use the real-world HM3DSem~\cite{yadav2023habitat} dataset, including 145 training scenes and 36 test scenes with semantic labels. From these scenarios, we collected over 200 different categories of targets based on the Goat-Bench~\cite{khanna2024goat} dataset. Then, different regions and their corresponding objects are fed into the Large Language Model (LLM), which combines them to generate task instructions and goals.

\noindent \textbf{Exploration Trajectory Generation.} Based on the agent's initial position and the positions of different targets, we used the Habitat-Sim~\cite{savva2019habitat} to plan the exploration path, thereby generating a multi-goal, step-by-step exploration sequence trajectory, including the corresponding action, observation, position, and rotation for each step. This is beneficial for the model to learn action-based step-by-step exploration in long-horizon planning. Simultaneously, we utilize an image tagging model~\cite{zhang2024recognize} to label object information in each image as a text description, thereby constructing a multi-modal memory bank including text, position, and image.

\noindent \textbf{Memory Bank.} To facilitate efficient data sampling, we construct a memory bank:
\begin{equation}
    \mathcal{M}=\{(p_i,f_i,o_i)\mid i=1,\ldots,n\},
\end{equation}
where each entry stores the position $p_i$, the text features $f_i$, and the image features $o_i$ for each step $i$. We employ CLIP~\cite{radford2021learning} to extract both $o$ and $f$, and compute their pairwise similarity using dot products. The overall similarity between the current state $(p_c, f_c, o_c)$ and a memory entry $(p_i, f_i, o_i)$ is defined as:
\begin{equation}
s_i = \omega_f (f_c^\top f_i) + \omega_o (o_c^\top o_i) + \omega_p \, \mathrm{dist}(p_c, p_i),
\end{equation}
where $\omega_f$, $\omega_o$, and $\omega_p$ are the weighting coefficients for text, visual, and distance similarities, respectively, and $\mathrm{dist}(p_c, p_i)$ denotes an exponential function of the Euclidean distance between locations. To maintain temporal consistency, we further aggregate similarity scores from the $k$ most recent samples by computing their mean and standard deviation, and dynamically filter contextual memories at each step based on an adaptive similarity threshold.

\begin{figure*}[!t]
\centering
\includegraphics[width=0.92\linewidth]{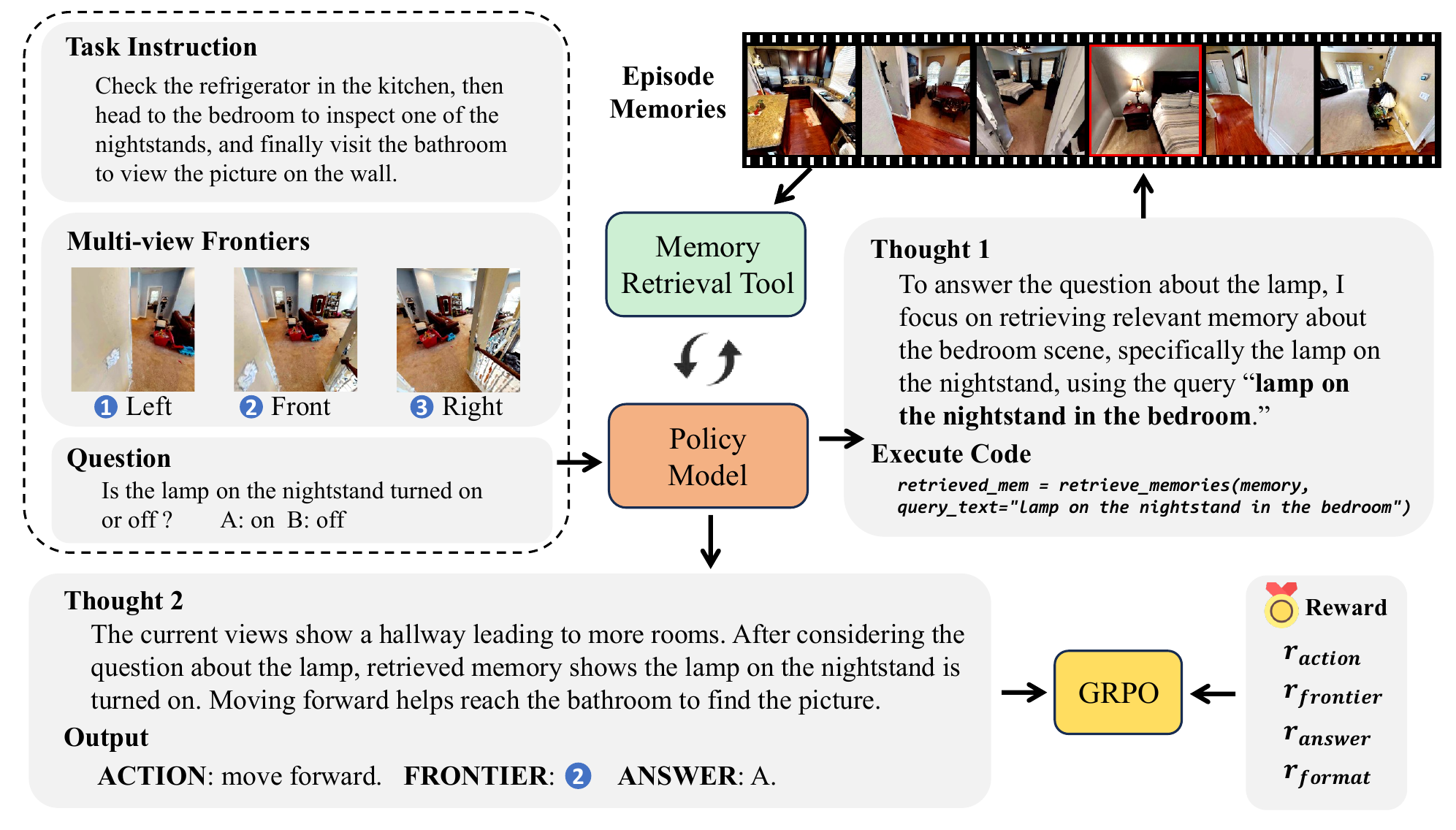}
\vspace{-0.6em}
\caption{Illustration of training in MemoryExplorer. Given a task instruction, the multi-view observations, and a goal-oriented question. Model retrieves relevant multimodal memories from the episodic memory bank using tools, analyzes the current information alongside the retrieved memories to understand the progress of the long-term task, and performs \texttt{ACTION} prediction, \texttt{FRONTIER} selection, and question \texttt{ANSWER}. The policy model output response calculates the reward using a Multi-Task Reward function and is fine-tuned using GRPO. }
\label{fig:method}
\vspace{-1.2em}
\end{figure*}

\noindent \textbf{Difficulty Level.} To evaluate the model’s exploration capabilities and long-term memory, we categorize tasks into three levels: easy, medium, and difficult, based on the number of regions and goals to be explored and the distance from the initial position to the target object. Detailed statistics are presented in \cref{fig:data_1}.

\noindent \textbf{Question Answering Generation.} We leverage a Vision Language Model (VLM) to generate question-answer pairs based on observation images associated with the navigation targets. This design serves two main purposes. First, questions are focused on navigation targets to avoid asking about objects an agent may not have observed due to non-unique trajectories, thus providing a more reliable assessment of its memory. Second, goal-oriented questions help the model understand the progress of multi-goal long-term tasks during training, determine the next goal to be found, and thus plan actions and paths. As shown in \cref{fig:data_1}, the types of questions include five categories: attribute, counting, location, relationship, and state. The types of answers include two forms: open-ended and optional.

\noindent \textbf{Continuous Actions.} We found that in many cases, the simulator cannot execute multi-step continuous actions, which greatly increases the difficulty of the model predicting actions. Therefore, we utilize a continuous action window to sample $x$ consecutively occurring identical actions, using one of such samples as training data.

\noindent \textbf{Dataset details.}
We generate step-by-step trajectory data using the training and testing scenes from HM3DSem, which serve as the training and testing sets, respectively, as shown in \cref{tab:intro_1}. We use Qwen3-235B-A22B-Instruct to generate task instructions and Qwen3-VL-235B-A22B-Instruct to generate question answering. The model’s action space includes moving forward (0.25 m) and turning left or right (30°), so each step contains views from these three directions. The full dataset comprises 1,982 tasks with a total of 377,311 entries, of which 1,816 tasks are allocated for training and 166 tasks for testing. After sampling, we obtain 11,684 instances for training.
For more details, please refer to the \textit{Supplementary Material}.
% We then construct an exploration memory bank from the trajectory data and sample multimodal data, including actions, observations, and question-answer pairs, to build the training set. $\omega_f$, $\omega_o$, and $\omega_p$ are 0.3, 0.5, 0.2, respectively. Actions are sampled at intervals of 20 steps, while both the memory sampling interval and the sample filtering window are set to 10 steps. Each consecutive action window spans 6 steps, resulting in a total of 11,684 training samples. For more details, please refer to the \textit{Supplementary Material}.

\section{Method}
\label{sec:Method}

Unlike previous modular learning-based vision-language navigation~\cite{chang2023goat, khanna2024goat, zhu2025move}, we utilize an MLLM to train an end-to-end embodied agent with proactive exploration awareness. In long-term embodied exploration tasks, MLLM needs to integrate task instructions, observed images, and long-term memories for scene planning and decision-making. However, due to the limitations of the context window, the model cannot access all long-term memories at once, and supervision of long-term multi-step actions may weaken the agent's autonomous exploration ability. Therefore, we propose MemoryExplorer, an embodied exploration model based on reinforcement learning with memory retrieval as shown in \cref{fig:method}.

\noindent \textbf{Task Definition.} For LMEE, the model policy can be represented as $\pi_\theta(I, O, Q; M)$ of parameterized model weights $\theta$. Model inputs include task instructions $I$, current multi-view observations $O$, a memory-based goal-oriented question $Q$, and externally stored long-range memory $M$ including image-text pairs. The model's final response can be represented as $y=(S, F, A)$, with outputs including a single-step action $S$, a frontier $F$, and an answer $A$.

\noindent \textbf{Memory Retrieval.} The proposed Long-term Memory Embodied Exploration dataset provides the multimodal contextual memories during the exploration process. The model is guided to generate code that invokes the external memory retrieval tool $\mathcal{R}$, subsequently obtaining the corresponding memories for reasoning and answering the given question. 
The memories obtained from invoking $\mathcal{R}$ are fed back to the model as additional input, allowing for richer reasoning to support the final answer. 
Due to the limitations of multi-image input, this study focuses on single-round tool invocation scenarios. Since the question is related to the navigation goal, the retrieved memories further help the model understand the task progress and determine the next exploration action. 

Formally, when the model retrieves memories, the first round generates an initial response $y' \sim \pi_\theta(\cdot \mid I, O, Q)$ containing the tool call and query text. Then, the memory retrieval tool $M_{re} = \mathcal{R}(y', M)$ is called to obtain the most relevant memories retrieved from long-term memory. In the external Python environment, the text and observations in memory are encoded as text features $f_t$ and observation features $f_o$ by CLIP~\cite{radford2021learning}, respectively, while the query text is encoded as query features $f_q$. Cosine similarity is then calculated to obtain the $topk$ most similar memories. Finally, the retrieved memories $m_i$ are obtained by combining the results from both. The retrieval process can be represented as:
\begin{equation}
\mathcal{R} = \{ m_i \mid i \in \mathrm{top}\text{-}k(\cos(f_q, f_i^{(t,o)})) \}.
\end{equation}
Finally, the final response can be represented as $y \sim \pi_\theta(\cdot \mid I, O, M_{re})$. Additionally, if the memory retrieval tool call fails, the final response is the first-round response, \ie{,} $y \sim \pi_\theta(\cdot \mid I, O, Q)$.

\noindent \textbf{Reinforcement Fine-Tuning.} MemoryExplorer utilizes RFT to learn active exploration and memory retrieval. Therefore, our training objective is:
\begin{equation}
\begin{aligned}
& \max_{\pi_\theta} \quad 
\mathbb{E}_{(I, O, Q) \sim D,\; y \sim \pi_\theta(\cdot \mid I, O, Q; M)} \left[r_\phi(I, O, Q, y)\right] \\
& \quad - 
\beta \,
D_{\mathrm{KL}}\!\left(
    \pi_\theta(\cdot \mid I, O, Q; M)
    \;\middle\|\;
    \pi_{\mathrm{ref}}(\cdot \mid I, O, Q; M)
\right).
\end{aligned}
\end{equation}
Note that our RFT training process does not optimize intermediate responses from memory retrieval tools, as our goal is to encourage the model to think and make decisions autonomously, using final reward feedback to evaluate the effectiveness of tool calls. For the specific training method, we employ Group Relative Policy Optimization (GRPO)~\cite{guo2025deepseek}, a well-established policy gradient method that estimates the baseline model by sampling response data, thus saving training resources.

\noindent \textbf{Reward Modeling.}
Unlike single-question-answer training~\cite{wu2025mmsearch, fu2025refocus, wu2025vtool}, we adopt a multi-task training approach that includes exploration actions, frontier selection, and memory-based question answering. This design enables the model to understand spatial and action relationships while actively invoking memory retrieval tools and generating retrieval content, thereby facilitating autonomous exploration.

To achieve multi-dimensional response optimization, we design a Multi-Task Reward function $r_{\text{total}}$ that integrates four complementary components: action accuracy, frontier correctness, answer precision, and output format completeness. 
The total reward is defined as follows:

\begin{equation}
\begin{aligned}
r_{\text{total}} =
& w_{act} \cdot r_{\text{action}} \cdot c +
w_{front} \cdot r_{\text{frontier}} \cdot c \\
& + w_{ans} \cdot r_{\text{answer}} +
w_{fmt} \cdot r_{\text{format}} \text{,}
\label{eq:reward_total}
\end{aligned}
\end{equation}
where $r_{\text{action}}, r_{\text{frontier}}, r_{\text{answer}}, r_{\text{format}} \in [0,1]$ represent the sub-rewards for action accuracy, frontier correctness, answer precision, and output format completeness, respectively. 
Specifically, $r_{\text{answer}}$ reflects the accuracy of the predicted answer, and $r_{\text{format}}$ encourages structured and parsable responses, determined by whether the output includes 
complete \texttt{ACTION}, \texttt{FRONTIER}, and \texttt{ANSWER} segments. $c$ is a consistency coefficient that penalizes logically inconsistent pairs between action and frontier. $w_{act}, w_{front}, w_{ans}, w_{fmt}$ denote the weighting coefficients for each reward component. 

To further differentiate performance in scenarios involving tool assistance and tool invocation failure, a scaling factor $\alpha$ is applied to each sub-reward. This adjustment reduces all sub-scores when no external tool is employed, while amplifying them in tool-based reasoning conditions, thereby encouraging efficient tool utilization. The final reward $r_{\text{total}}$ is clipped to the range $[0,1]$ to ensure stability and comparability across tasks.
\section{Experiments}
\label{sec:experiment}

\begin{table*}[!t]
  \centering
  \caption{Experiments on LMEE-Bench. Score represents the MLLM-Score for open-ended answers, and Acc represents the accuracy rate of the answer choices.}
  \vspace{-0.6em}
    \renewcommand{\arraystretch}{1.2}
   \resizebox{0.96\linewidth}{!}{
    \begin{tabular}{c|cc|cc|cc|cc|cc|cc|cc|cc}
    \toprule
    \multirow{3}[6]{*}{Method} & \multicolumn{4}{c|}{Easy}     & \multicolumn{4}{c|}{Medium}   & \multicolumn{4}{c|}{Hard}     & \multicolumn{4}{c}{Total} \\
\cmidrule{2-17}          & \multicolumn{2}{c|}{Nav} & \multicolumn{2}{c|}{QA} & \multicolumn{2}{c|}{Nav} & \multicolumn{2}{c|}{QA} & \multicolumn{2}{c|}{Nav} & \multicolumn{2}{c|}{QA} & \multicolumn{2}{c|}{Nav} & \multicolumn{2}{c}{QA} \\
\cmidrule{2-17}          & SR    & SPL   & Score & Acc   & SR    & SPL   & Score & Acc   & SR    & SPL   & Score & Acc   & SR    & SPL   & Score & Acc \\
    \midrule
\textit{\textbf{Retrieval-Augmented QA}}          &       &       &       &       &       &       &       &       &       &       &       &       &       &       &       &  \\
LLaVA-OneVision-7B~\cite{li2024llava} &- &- &35.71   & 57.14  &- &- &42.02  & 62.77 &-  &-  &25.00   & 43.75 &- &- & 38.62 &59.31  \\
Llama3.2-Vision-11B~\cite{meta2024llama} &- &- &43.57  & 48.57 &-  &-  &28.19  &37.23  & - & - &20.31   & 31.25  &- &- & 31.03  & 39.31 \\
Qwen2.5-VL-7B~\cite{bai2025qwen2} &-  &- &36.43   & 65.71 &-   &-  & 43.62 &53.19   &-    &-   & 42.19 & 43.75   & - & - & 41.72 & 55.86 \\
InternVL3-8B~\cite{zhu2025internvl3}  &-  &- &42.14  & 65.71 &-   &-  & 41.22 &54.26   &-    &-   & 26.56 & 43.75   & - & - & 39.83 & 55.17 \\
Qwen3-VL-8B~\cite{team2025qwen3} & - & - & 43.57 & 65.71 &-  &-  & 36.70 & 58.51  & - & - & 32.81 & 62.50 & - & - & 37.93   & 60.69 \\
GPT-4o & - & - & 43.57 & 65.71 &-  &-  & 37.50 & 51.06  & - & - & 40.62 & 56.25 & - & - & 39.31   & 55.17 \\
Gemini-2.5-Flash & - & - & 44.29 & 51.43 &-  &-  & 37.77 & 50.00  & - & - & 23.44 & 50.00 & - & - & 37.76   & 50.34 \\
          \midrule
          \textit{\textbf{MLLM Exploration}} &       &       &       &       &       &       &       &       &       &       &       &       &       &       &       &  \\
         Explore-EQA~\cite{ren2024explore} & 4.26  & 4.26  &- &-  & 15.17 & 8.67  &-  &-  & 14.89 & 7.25  & -  & - & 13.24 & 7.66  & -  & - \\
   3D-Mem~\cite{yang20253d} & 21.28  &  9.68 &   30.71 & 42.86 & 15.73 &  6.09  &  34.57  & 44.68 &  17.02 &  6.97 & 25.00  &18.75   &16.91 &6.86 &32.59 &41.38  \\
    RA-Mem & 25.53  & 15.65 &  34.29 & 54.29 & \textbf{21.35} & 12.14 & 37.77 &62.77 & 14.89 &8.86  & 25.00   & 43.75  & 20.96   &  12.18  & 35.52      & 58.62 \\
    \rowcolor{blue!10}
    MemoryExplorer (Ours)  &\textbf{31.91} &\textbf{21.11} & \textbf{35.71}  & \textbf{68.57} & \textbf{21.35}  & \textbf{14.03}    & \textbf{48.14}  & \textbf{63.83}  & \textbf{23.40}  & \textbf{12.51} & \textbf{34.38} & \textbf{68.75}  &\textbf{23.53}  &\textbf{14.99} & \textbf{43.62} & \textbf{65.52} \\
    \bottomrule
    \end{tabular}%
    }
    \vspace{-0.6em}
  \label{tab:ex_1}%
\end{table*}%

We propose the Long-term Memory Embodied Exploration benchmark. It consists of two components: multi-goal navigation and memory-based question answering. The agent is first required to perform multi-goal navigation in an unknown environment, storing and utilizing memories from the exploration process to complete the tasks. Subsequently, the agent must answer questions related to the navigation targets based on its memory. Additionally, during multi-goal navigation, the agent can leverage memory retrieval to locate specified targets. We also evaluate the agent’s exploration and memory retrieval capabilities on the multimodal lifelong navigation benchmark platform, GOAT-Bench.

\noindent \textbf{Experimental details.}
Our model is trained based on Qwen2.5-VL-7B-Instruct~\cite{bai2025qwen2}. We use EasyR1, a simplified version of the VERL framework. The learning rate is set to 1e-6, and a KL penalty coefficient of 0.1 is applied to maintain training stability. Training is conducted on 8 NVIDIA H200 GPUs for 160 steps with a global batch size of 128. The $topk$ is set to 3. The consistency coefficient $c$ is set to 0.5. The scaling factor $\alpha$ is set to 1.2 when tool assistance is involved, and to 0.5 for $r_{\text{answer}}$ and $r_{\text{format}}$, and 0.6 for $r_{\text{action}}$ and $r_{\text{frontier}}$ when tool invocation fails. More experimental details are provided in the \textit{Supplementary Material}.

\begin{figure*}[!t]
\centering
\includegraphics[width=16.5cm, height=8cm]{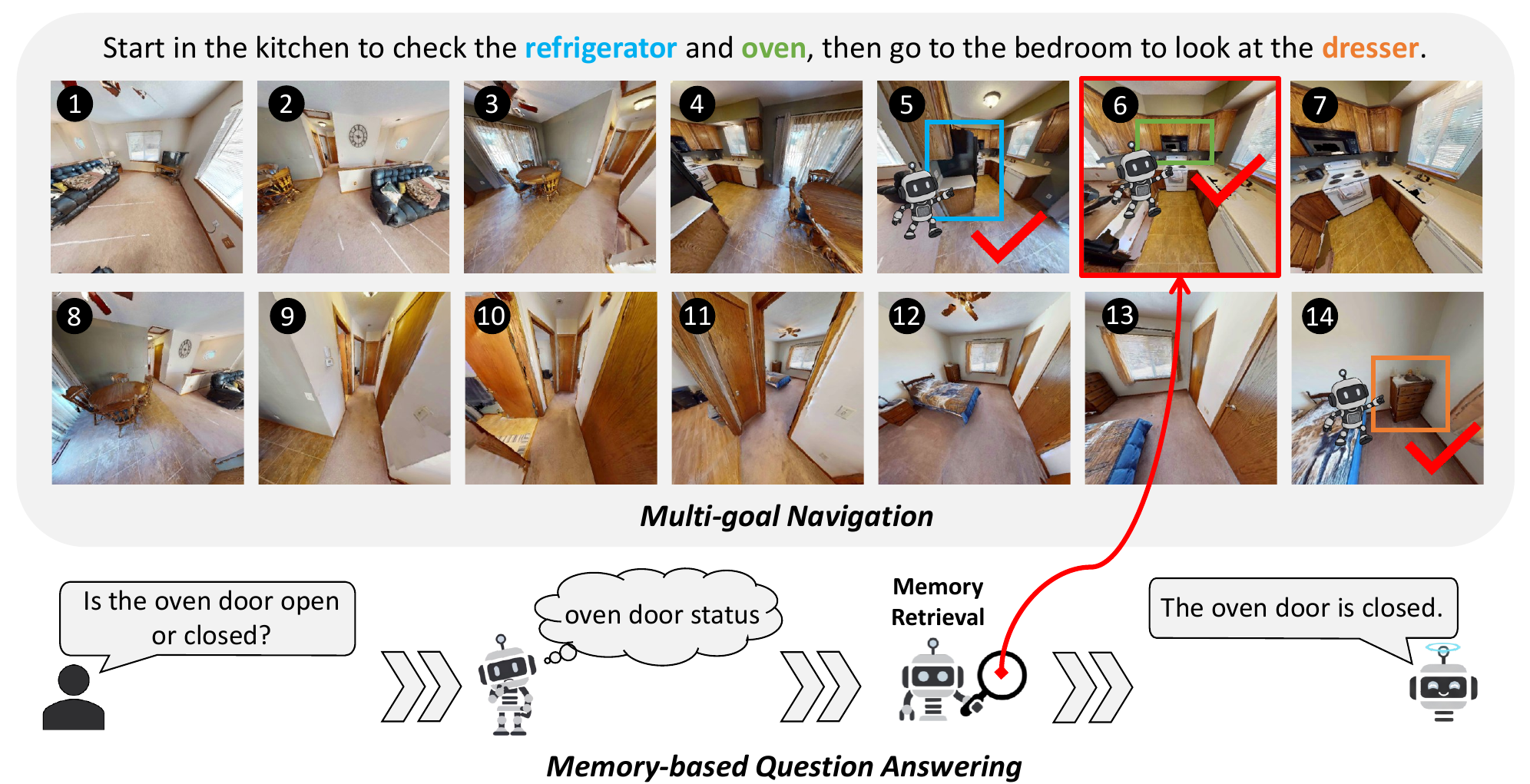}
\vspace{-0.6em}
\caption{Qualitative example of LMEE-Bench.}
\label{fig:case}
\vspace{-1.2em}
\end{figure*}

% \subsection{Long-term Memory Embodied Exploration}
% \label{sec:lmee}
\noindent \textbf{Benchmark.} \textbf{LMEE-Bench} consists of 166 tasks with 828 goals and 406 questions. Due to resource constraints, we randomly select a subset of 58 tasks for evaluation, covering 272 goals and 145 questions. We also provide results on the full test set in the \textit{Supplementary Material}. LMEE consists of two types of tasks: multi-goal navigation and goal-oriented memory-based question answering. The questions cover five types: attributes, counting, location, relationships, and states, while the answers are provided in two formats: open-ended and multiple-choice. \textbf{GOAT-Bench} is a multimodal lifelong navigation benchmark. The task requires agents to navigate to multiple targets. Each target is described by category name, language description, and image. Due to the large scale and limited resources of GOAT-Bench, we followed 3D-Mem to evaluate a subset of ``Val Unseen" set, including 36 scenarios, one exploration round each, and 278 navigation subtasks in total.

\noindent \textbf{Metric.}
We evaluate performance using Success Rate (SR) and Success weighted by Path Length (SPL). A navigation episode is considered successful if the agent's final position is within 1 meter of the target. For goal-oriented open-ended questions, we propose MLLM-Score, a quantitative metric leveraging an MLLM. For each question, the ground-truth answer, the observation of the target object, and the predicted answer are provided to the MLLM. Due to resource constraints, we use Qwen3-VL-30B-A3B-Instruct as our evaluation model. The MLLM assigns a score from 1 to 5 to each prediction to assess its quality. The MLLM-Score measures answer accuracy by averaging the scores across all questions and converting the result to a 0–100 scale. For multiple-choice questions, we evaluate predicted answers using standard accuracy.

\noindent \textbf{Baselines.}
For the baseline evaluation of long-term memory exploration with MLLMs, we primarily compare our method with Explore-EQA~\cite{ren2024explore} and 3D-Mem~\cite{yang20253d}. Since Explore-EQA lacks memory capabilities, we only evaluate its performance on multi-goal navigation tasks. 3D-Mem serves as an active exploration baseline leveraging long-term memory, enabling agent exploration and lifelong learning through memory snapshots and frontier snapshots.
In addition, we develop a Retrieval-Augmented Memory (RA-Mem) variant based on 3D-Mem. While 3D-Mem uses an object-based memory filtering approach to limit the model’s context window, it does not fully exploit the potential of MLLMs. In contrast, RA-Mem independently generates queries based on the task and current observations to retrieve memory and guide the agent in completing tasks. However, these methods still rely solely on MLLM reasoning.
Our proposed MemoryExplorer extends RA-Mem by incorporating reinforcement learning, improving the model’s active memory retrieval and exploration capabilities. Beyond active exploration, we also evaluate a post-exploration setting, where MemoryExplorer collects observations during navigation to construct a memory bank, and we compare the performance of different MLLM models using Retrieval-Augmented Question Answering.

\begin{table}[!t]
  \centering
  \caption{Experiments on GOAT-Bench. Evaluated on the ``Val
Unseen" split. Methods denoted by * are from GOAT-Bench, and those with † are evaluated on the subset. All MLLM-based exploration methods are implemented based on Qwen2.5-VL-7B.}
\vspace{-0.6em}
    \renewcommand{\arraystretch}{1.2}
 \resizebox{0.8\linewidth}{!}{
    \begin{tabular}{lcc}
    \toprule
    Method & \multicolumn{1}{c}{Success Rate} & \multicolumn{1}{c}{SPL} \\
    \midrule
    \textit{\textbf{GOAT-Bench Baselines}}~\cite{khanna2024goat} &       &  \\
    Modular GOAT$^*$   & 24.9   & 17.2 \\
    Modular CLIP on Wheels$^*$  & 16.1  & 10.4  \\
    SenseAct-NN Skill Chain$^*$ & 29.5   & 11.3 \\
    SenseAct-NN Monolithic$^*$ & 12.3  & 6.8 \\
    \midrule
    \textit{\textbf{MLLM Exploration}} &       &  \\
    Explore-EQA~\cite{ren2024explore}$^\dagger$      & 23.02  & 14.43 \\
    3D-Mem~\cite{yang20253d}$^\dagger$      & 37.05   & 20.26 \\
    RA-Mem$^\dagger$      & 42.81   & 21.95 \\
    \rowcolor{blue!10}
    MemoryExplorer (Ours)$^\dagger$  & \textbf{46.40}  & \textbf{28.03} \\
    \bottomrule
    \end{tabular}%
    }
    \vspace{-1.8em}
  \label{tab:ex_2}%
\end{table}%

\noindent \textbf{Quantitative Comparison.}
As shown in \cref{tab:ex_1}, our model demonstrates higher robustness and efficiency compared to existing embodied exploration methods on LMEE-Bench. In addition, we evaluate current MLLMs using retrieval-augmented question answering. We observe that Qwen2.5-VL-7B performs better on open-ended questions, while Qwen3-VL-8B and LLAVA-OV-7B excel at multiple-choice questions. This suggests that misalignment between cognitive understanding and action decision-making may lead to hallucinations in the models. 

The results on GOAT-Bench are shown in \cref{tab:ex_2}, RA-Mem is more flexible than the memory pre-filtering mechanism used in 3D-Mem, as it actively generates memory query texts to more effectively leverage long-term memory, thereby significantly improving the model’s success rate. Moreover, MemoryExplorer further enhances the agent’s success rate and efficiency in multi-goal long-horizon navigation, demonstrating the performance gains brought by improved utilization of exploratory memories.

\noindent \textbf{Qualitative results on LMEE-Bench.}
To intuitively demonstrate the effectiveness of long-term memory–based embodied exploration, we present a test case on LMEE-Bench in \cref{fig:case}. Through multi-goal exploration and memory accumulation, the agent progressively understands the environment. When a question is presented, the agent retrieves relevant memories and answers the question.

\noindent \textbf{Ablation Study.} Our baseline is RA-Mem, and we apply RFT to enhance the agent’s ability to actively retrieve memories and exploration. We begin with a relatively simple task-progress question: “\textit{Which objects in the task have already been found or completed?}” The answer corresponds to the discovered target objects, such as “\textit{tv, couch, oven, sink.}”
We then increase the difficulty by extending it to multiple-choice questions covering all five dataset question types. Finally, we combine the two settings above. As shown in \cref{tab:ab_1}, training the model using memory retrieval tools significantly improved model performance, while there is a non-linear positive correlation between the question types and model performance. Using only a single question type cannot achieve the best results, whereas incorporating a richer variety of question types enables the model to achieve strong performance.

\begin{figure}[htbp]
\centering
\vspace{-0.6em}
\includegraphics[width=0.98\linewidth]{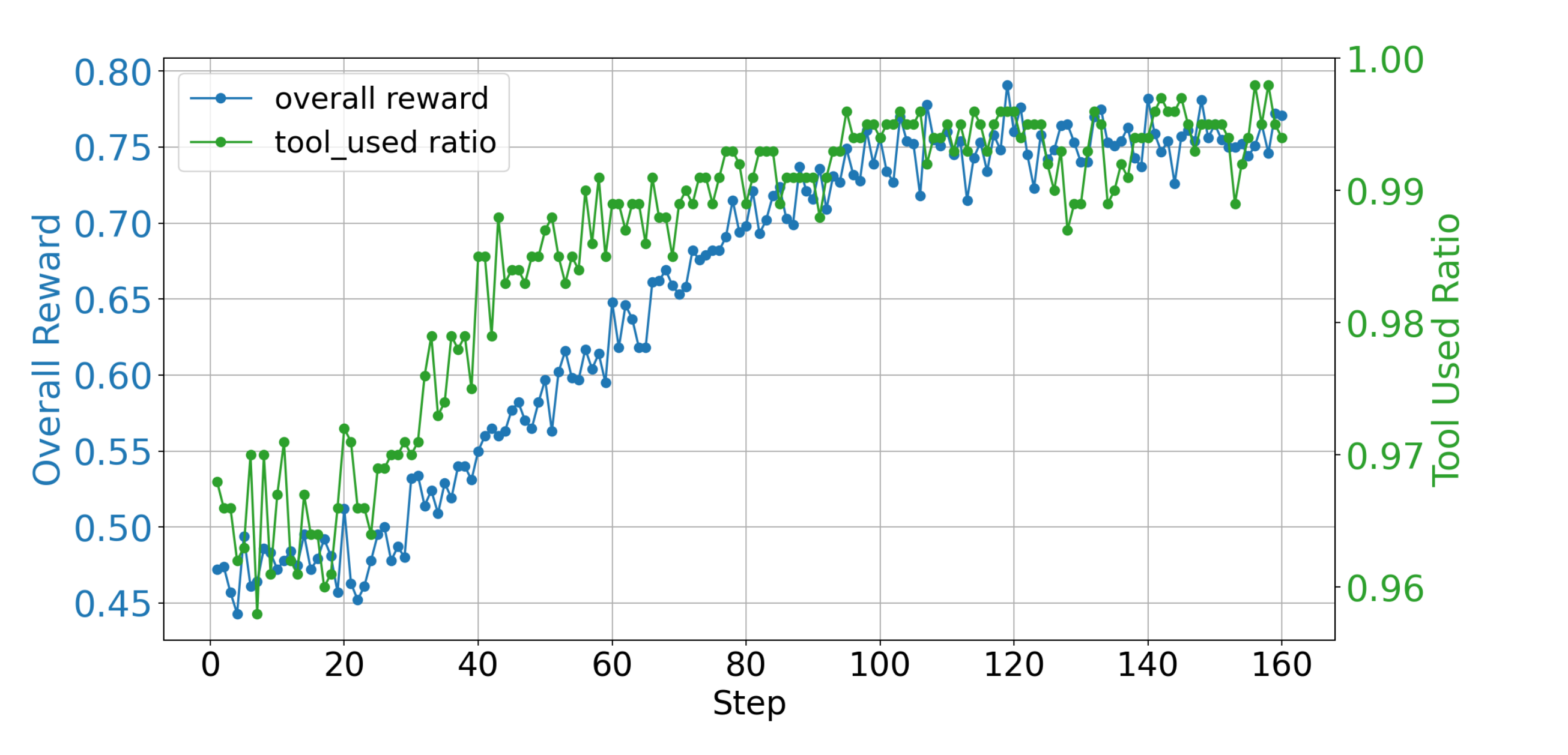}
\caption{Training reward curve and tool usage percentage.}
\label{fig:ex_4}
\vspace{-1.2em}
\end{figure}

\noindent \textbf{Visualization.}
We visualize the training reward curve and tool usage percentage in \cref{fig:ex_4}.The model gradually learns to invoke the memory-retrieval tool more accurately, which in turn leads to improved answer accuracy.

\begin{table}[!t]
  \centering
  \caption{Ablation study on question type.}
      \vspace{-0.6em}
      \renewcommand{\arraystretch}{1.2}
       \resizebox{0.9\linewidth}{!}{
    \begin{tabular}{c|ccc|cc}
    \toprule
    \multirow{2}[4]{*}{Setting} & \multicolumn{3}{c|}{LMEE-Bench} & \multicolumn{2}{c}{GOAT-Bench} \\
\cmidrule{2-6}          & SR    & SPL   & Score & SR    & \multicolumn{1}{c}{SPL} \\
    \midrule
    Baseline &20.96 &12.18  &35.52 & 42.81 & 21.95 \\
    Simple &20.80  &12.49 &41.33 &44.24  &27.29  \\
    Multiple-choice &\textbf{23.53}   &14.99 &\textbf{43.62}  &46.40  &28.03  \\
    All &22.06 &\textbf{15.13} &43.28 &\textbf{48.20}  &\textbf{29.36}\\
    \bottomrule
    \end{tabular}%
    }
  \label{tab:ab_1}%
    \vspace{-1.2em}
\end{table}%

\section{Conclusion}
\label{sec:conclusion}
We propose Long-term Memory Embodied Exploration (LMEE), which constructs a episodic memory bank through Multi-goal Navigation and leverages Memory-based Question Answering to promote the integration of cognitive  and decision-making. We further introduce MemoryExplorer, a reinforcement learning framework that trains the model to actively retrieve memory. A Multi-Task reward function combining action prediction, frontier selection, and question answering enables autonomous exploration and proactive memory use. Our approach achieves strong performance on both the LMEE-Bench and GOAT-Bench.

\section*{Acknowledgments}
This work was supported by the National Natural Science Foundation of China (Grant Nos. 62302167, 62222602, 62502159, U23A20343, W2521174), the Shanghai Committee of Science and Technology (Grant Nos. 25511103300, 25511104302, 25511102700), the Natural Science Foundation of Chongqing (CSTB2023NSCQ-JQX0007), and the Young Elite Scientists Sponsorship Program by CAST YESS20240780.
{
    \small
    \bibliographystyle{ieeenat_fullname}
    \bibliography{main}
}
\clearpage
\setcounter{page}{1}
\maketitlesupplementary

\section{Full-set Evaluation}
Due to resource constraints, we used approximately 35\% of the test set (58/166) for comparison in the main text. To fully illustrate the superiority of our method, we present the comparison results of existing embodied exploration methods on the full LMEE-Bench in \cref{tab:full}. Due to the limitation of inference speed, completing all 166 tasks requires a substantial amount of time. The experimental conclusions show no significant difference between the subset and the full test set, demonstrating that the subset is sufficient for accurately evaluating the model.

\begin{figure}[htbp]
\centering
\includegraphics[width=0.98\linewidth]{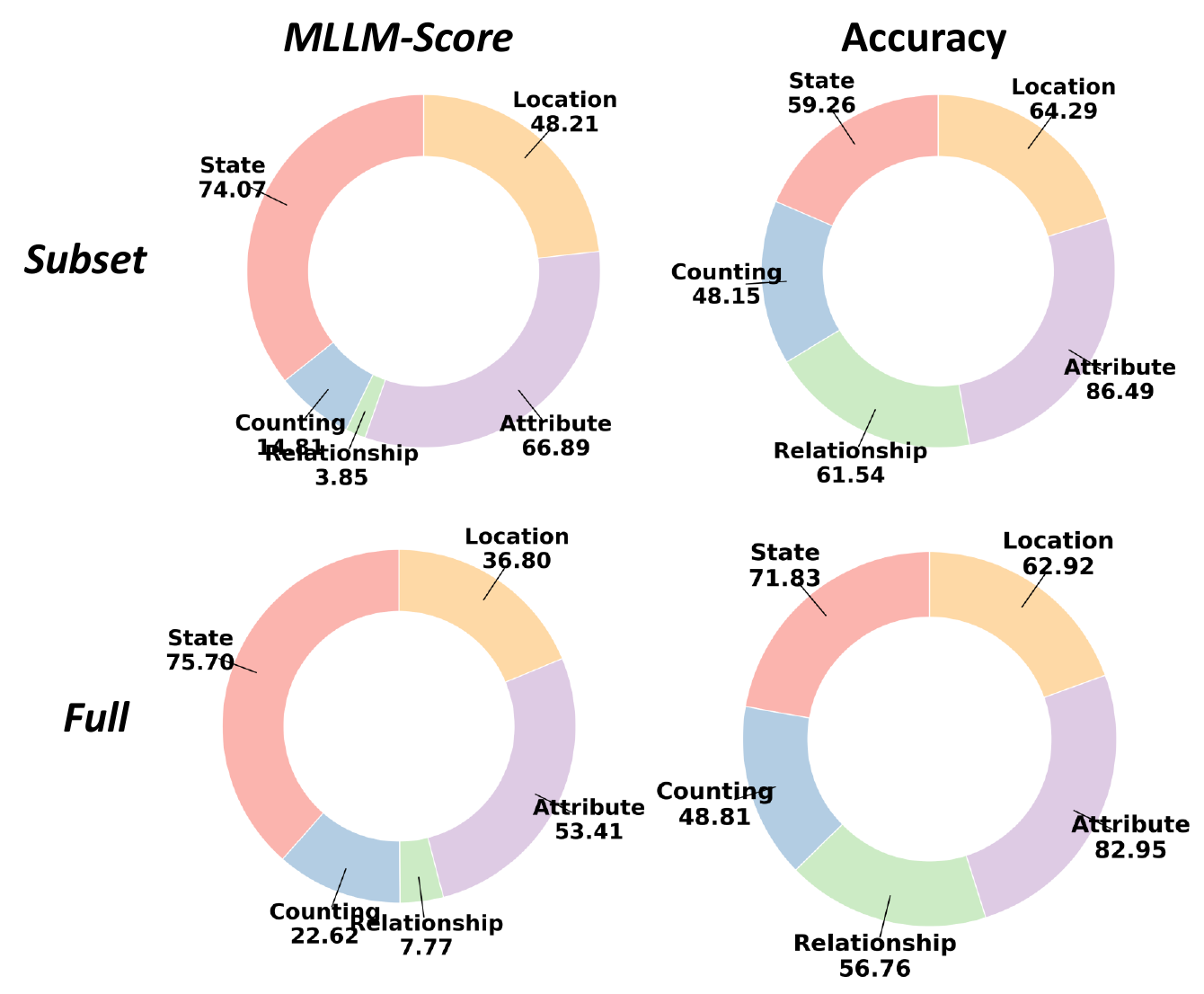}
\caption{Answer quality across different question types.}
\label{fig:answer}
\end{figure}

We also evaluate the answer quality across different question types. As shown in \cref{fig:answer}, we test MemoryExplore on both the subset and the full dataset, and the distribution of performance across question types remains largely consistent, further demonstrating the generalizability of the subset results. In addition, for counting and relational questions, there is a noticeable performance gap between open-ended answers and multiple-choice answers, indicating that large models still struggle with certain challenging open-ended question types.

\begin{table*}[!t]
  \centering
  \caption{Experiments on subset and full-set LMEE-Bench.}
  \resizebox{0.98\linewidth}{!}{
    \begin{tabular}{c|cc|cc|cc|cc|cc|cc|cc|cc|r}
    \toprule
    \multirow{3}[6]{*}{Method} & \multicolumn{4}{c|}{Easy}     & \multicolumn{4}{c|}{Medium}   & \multicolumn{4}{c|}{Hard}     & \multicolumn{4}{c|}{Total}    & \multicolumn{1}{c}{\multirow{3}[6]{*}{Time}} \\
\cmidrule{2-17}          & \multicolumn{2}{c|}{Nav} & \multicolumn{2}{c|}{QA} & \multicolumn{2}{c|}{Nav} & \multicolumn{2}{c|}{QA} & \multicolumn{2}{c|}{Nav} & \multicolumn{2}{c|}{QA} & \multicolumn{2}{c|}{Nav} & \multicolumn{2}{c|}{QA} &  \\
\cmidrule{2-17}          & SR    & SPL   & Score & Acc   & SR    & SPL   & Score & Acc   & SR    & SPL   & Score & Acc   & SR    & SPL   & Score & Acc   &  \\
    \midrule
    \textit{\textbf{Subset}}      &       &       &       &       &       &       &       &       &       &       &       &       &       &       &       &       &  \\
         Explore-EQA~\cite{ren2024explore} & 4.26  & 4.26  &- &-  & 15.17 & 8.67  &-  &-  & 14.89 & 7.25  & -  & - & 13.24 & 7.66  & -  & -  &2h  \\
   3D-Mem~\cite{yang20253d} & 21.28  &  9.68 &   30.71 & 42.86 & 15.73 &  6.09  &  34.57  & 44.68 &  17.02 &  6.97 & 25.00  &18.75   &16.91 &6.86 &32.59 &41.38   & 17h  \\
    RA-Mem & 25.53  & 15.65 &  34.29 & 54.29 & \textbf{21.35} & 12.14 & 37.77 &62.77 & 14.89 &8.86  & 25.00   & 43.75  & 20.96   &  12.18  & 35.52      & 58.62  &10h  \\
    \rowcolor{blue!10}
    MemoryExplorer (Ours)  &\textbf{31.91} &\textbf{21.11} & \textbf{35.71}  & \textbf{68.57} & \textbf{21.35}  & \textbf{14.03}    & \textbf{48.14}  & \textbf{63.83}  & \textbf{23.40}  & \textbf{12.51} & \textbf{34.38} & \textbf{68.75}  &\textbf{23.53}  &\textbf{14.99} & \textbf{43.62} & \textbf{65.52}   & 10h \\ 
    \midrule
    \textit{\textbf{Full}}  &       &  &       &       &       &       &       &       &       &       &       &       &       &       &       &       &  \\
Explore-EQA~\cite{ren2024explore} &9.40 & 3.44   &-  &-   &10.02  &5.78  &-&- &7.59 &5.12&- &-  &9.50  &5.33  &- &- &8h  \\
3D-Mem~\cite{yang20253d} &29.06 &15.43  &30.94   &37.50  &16.10  &7.05  &32.35  &38.60  &14.48  &6.38 &22.69 &44.44   &17.66    &8.13 &30.79 &39.16  &49h  \\
RA-Mem &29.91 &16.66   &\textbf{40.62}  &53.75 &20.04  &11.10   &34.47   &56.25  &14.48  &8.51  &27.31 &51.85  &20.46   &11.44  &34.73  &55.17  &29h  \\
\rowcolor{blue!10}
MemoryExplorer (Ours) &\textbf{33.33}  &\textbf{20.62}   &37.50  &\textbf{66.25}   &\textbf{20.39} &\textbf{13.99} &\textbf{41.18} &\textbf{64.34}  &\textbf{19.31} &\textbf{10.21}  &\textbf{30.09}  &\textbf{64.81} &\textbf{22.05}  &\textbf{14.26}  & \textbf{38.98}  & \textbf{64.78} &29h  \\
    \bottomrule
    \end{tabular}%
    }
  \label{tab:full}%
\end{table*}%

\section{Ablation Study}
First, we provide additional experiments to verify the effectiveness of the training task design. As shown in \cref{tab:suppl_ab_1}, supervising the agent only on autonomous navigation does not lead to a clear performance improvement. Once the memory-retrieval tool is introduced, the model achieves a significant performance gain, demonstrating that learning active retrieval is crucial for improving accuracy in long-horizon navigation and memory-based question answering.

Second, we conduct ablation studies on the reward design. Our proposed multi-task reward consists of an action–frontier consistency penalty and a tool-usage penalty. As shown in \cref{tab:suppl_ab_2}, incorporating these penalties leads to improved model performance.

Finally, \cref{tab:suppl_ab_3} presents additional ablation experiments on the training hyperparameter settings.
\begin{table}[htbp]
  \centering
  \caption{Ablation study on training task setting.}
    \resizebox{0.8\linewidth}{!}{
    \begin{tabular}{c|cccc}
    \toprule
    \multirow{2}[4]{*}{} & \multicolumn{4}{c}{LMEE-Bench} \\
\cmidrule{2-5}          & SR    & SPL   & Score & Acc \\
    \midrule
    baseline & 20.96 & 12.18 & 35.52 & 58.62 \\
    only nav & 20.59 & 12.34 & 39.14 & 58.62 \\
    w/ memory tool & \textbf{23.53} & \textbf{14.99} & \textbf{43.62} & \textbf{65.52} \\
    \bottomrule
    \end{tabular}%
    }
  \label{tab:suppl_ab_1}%
\end{table}%

\begin{table}[htbp]
  \centering
  \caption{Ablation study on reward design.}
    \resizebox{0.9\linewidth}{!}{
    \begin{tabular}{c|cccc}
    \toprule
    \multirow{2}[4]{*}{} & \multicolumn{4}{c}{LMEE-Bench} \\
\cmidrule{2-5}          & SR    & SPL   & Score & Acc \\
    \midrule
    MemoryExplorer & \textbf{23.53} & \textbf{14.99} & \textbf{43.62} & \textbf{65.52} \\
    w/o consistency penalty $c$ & 22.43 & 13.76 & 41.38 & \textbf{65.52} \\
    w/o tool-usage penalty $\alpha$ & 21.32 & 12.79  & 41.03 & 64.14 \\
    \bottomrule
    \end{tabular}%
    }
  \label{tab:suppl_ab_2}%
\end{table}%

\begin{table}[htbp]
  \centering
  \caption{Ablation study on hyperparameters.}
  \resizebox{0.9\linewidth}{!}{
    \begin{tabular}{cccc|cccc}
    \toprule
    \multirow{2}[4]{*}{$w_{\text{act}}$} & \multirow{2}[4]{*}{$w_{\text{front}}$} & \multirow{2}[4]{*}{$w_{\text{ans}}$} & \multirow{2}[4]{*}{$w_{\text{fmt}}$} & \multicolumn{4}{c}{LMEE-Bench} \\
\cmidrule{5-8}          &       &       &       & SR    & SPL   & Score & Acc \\
    \midrule
    0.2   & 0.2   & 0.4   & 0.2   &\textbf{23.53} & \textbf{14.99} & \textbf{43.62} &65.52 \\
    0.3   & 0.3   & 0.3   & 0.1   &22.43 &14.65 &42.76 &\textbf{66.90} \\
    0.4   & 0.4   & 0.1   & 0.1   &22.43 &12.43 &42.24 &64.83 \\
    \bottomrule
    \end{tabular}%
    }
  \label{tab:suppl_ab_3}%
\end{table}%

\section{Real World Testing}

\begin{figure}[htbp]
\centering
\includegraphics[width=0.6\linewidth]{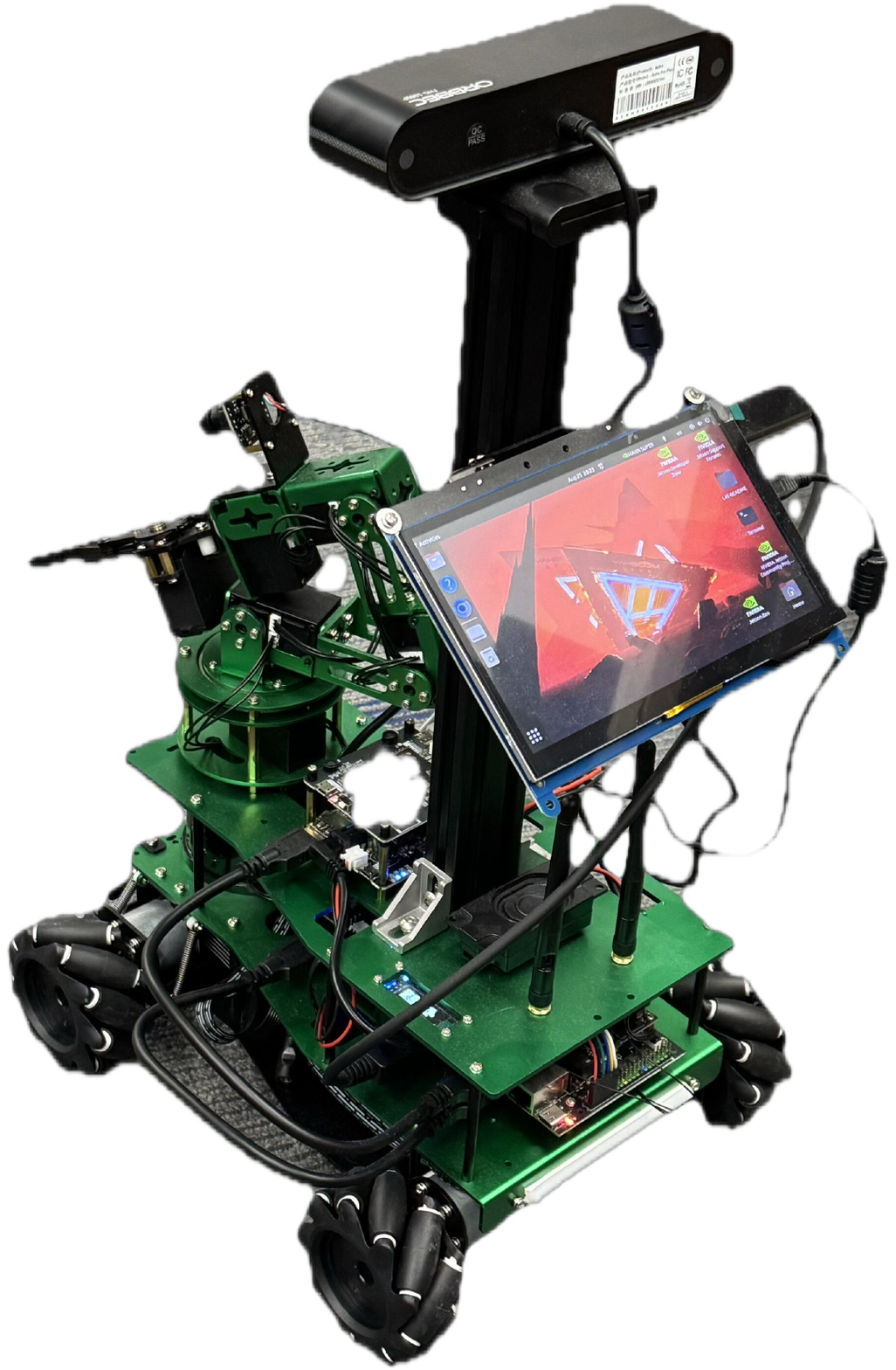}
% \vspace{-0.6em}
\caption{ROSMASTER X3.}
\label{fig:real_car}
% \vspace{-1.2em}
\end{figure}

To verify the sim-to-real generalization and practical applicability of our MemoryExplorer agent, we deployed it on a physical robotic platform. This section details our experimental setup and presents the qualitative results from tests conducted in real-world, unstructured office environments. Our goal was to demonstrate that the core long-term memory mechanism can be effectively transferred from simulation to reality. The experiments were performed on a ROSMASTER X3 robot as shown in \cref{fig:real_car}, which uses an Orbbec Astra Pro depth camera as its primary visual sensor. Our system architecture involved the robot, a local computer, and a remote server with an NVIDIA H200 GPU.

We evaluate the agent’s ability to perform multi-goal navigation and utilize its memory in two different office environments (a meeting room and a reception room).
In the first case, conducted in a meeting room, the task instruction is: “\textit{Start in the meeting room to find a bottle of water, then look for a rubbish bin.}” After completing the navigation task, we construct the memory bank from the observation images and then perform memory-based question answering. When asked “\textit{Where is the bottle of water?}”, MemoryExplorer retrieves the relevant memory and responds: “\textit{The bottle of water is on the chair.}”

In the second example, the task instruction is: “\textit{Start in the reception room to find a rubbish bin, a potted cactus, and an umbrella.}” After finishing navigation, we intentionally ask a question unrelated to the navigation targets: “Where is the tripod? Please describe its location in detail.” MemoryExplorer generates an appropriate retrieval query and accurately answers: “\textit{The tripod is in the corner of the room, near a wall with a blue triangle on it.}”

\begin{figure*}[!t]
\centering
\includegraphics[width=1\linewidth]{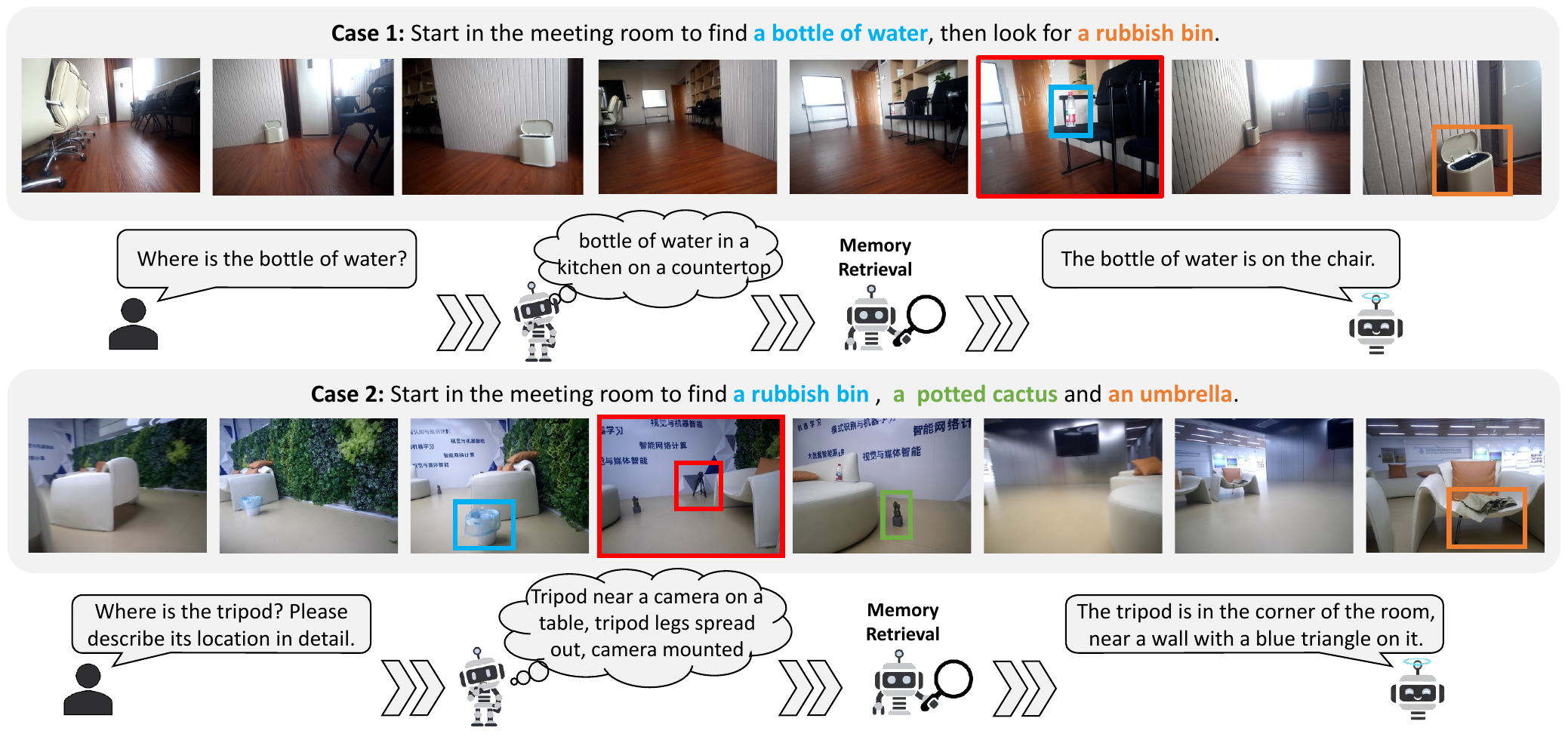}
% \vspace{-0.6em}
\caption{Real-world testing.}
\label{fig:real}
% \vspace{-1.2em}
\end{figure*}

In summary, our real-world experiments demonstrate the robustness and generalization ability of MemoryExplorer, highlighting the strong coupling between cognition and decision-making that lies at the core of our approach.

\section{Illustration of the Full Task Process}

\begin{figure*}[!t]
\centering
\includegraphics[width=1\linewidth]{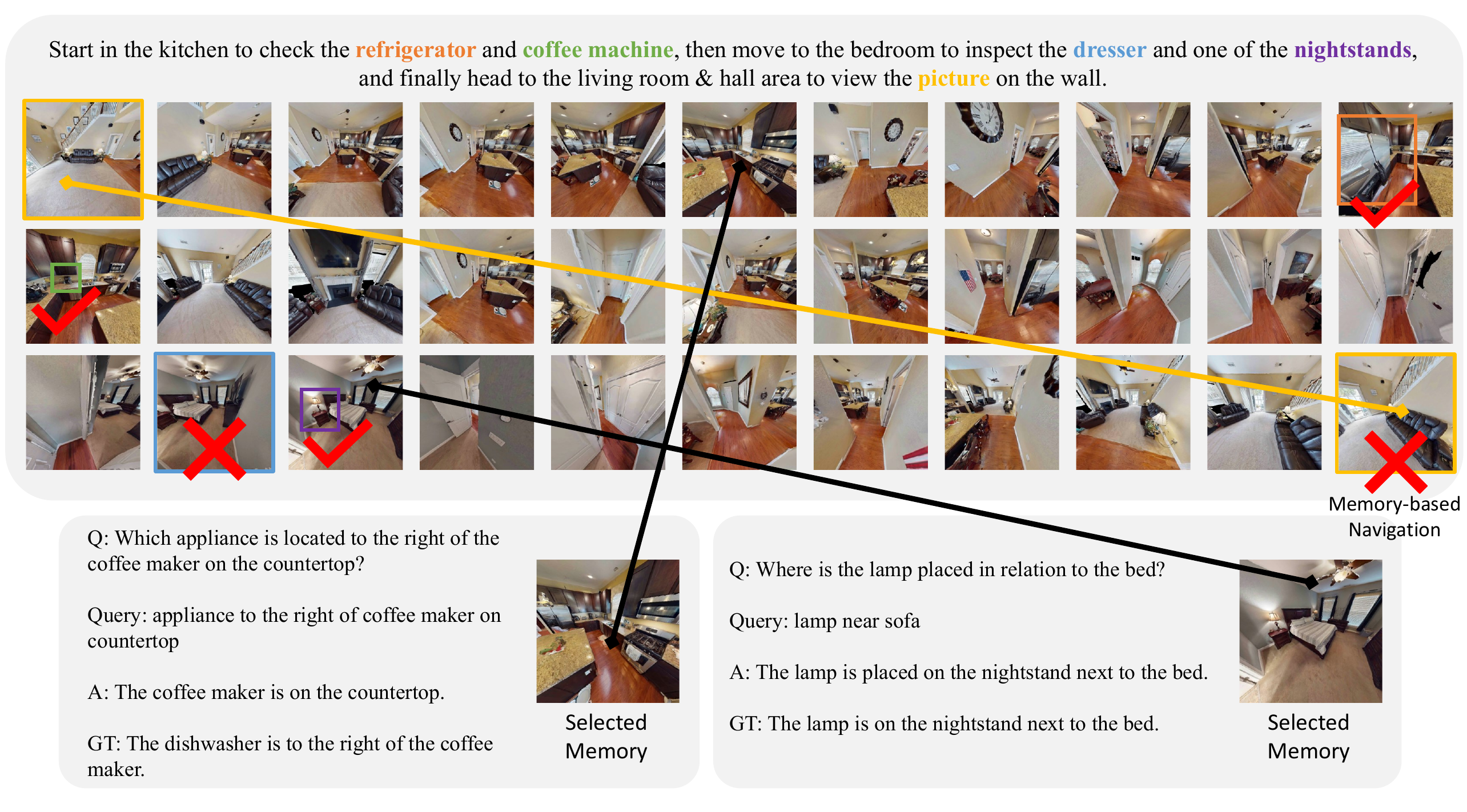}
% \vspace{-0.6em}
\caption{Complete task. Task is executed in a left-to-right, top-to-bottom order.}
\label{fig:complete}
% \vspace{-1.2em}
\end{figure*}

In \cref{fig:complete}, we present the content of a complete task, which consists of five navigation goals and two memory-based QA tasks. The agent successfully locates the refrigerator, coffee machine, and nightstand, but fails to find the dresser and picture due to incorrect memory retrieval. When searching for the picture, the agent retrieves the picture that appears in its memory, locates its position, and approaches it, but the target it is looking for is not the picture in its memory.
For the memory-based QA tasks, the first question about the coffee machine is answered incorrectly because the retrieved memory is inaccurate. In contrast, the second question selects the correct memory entry, enabling the agent to produce the correct answer.

\section{Failure Case Analysis}

\begin{figure}[!t]
\centering
\includegraphics[width=0.8\linewidth]{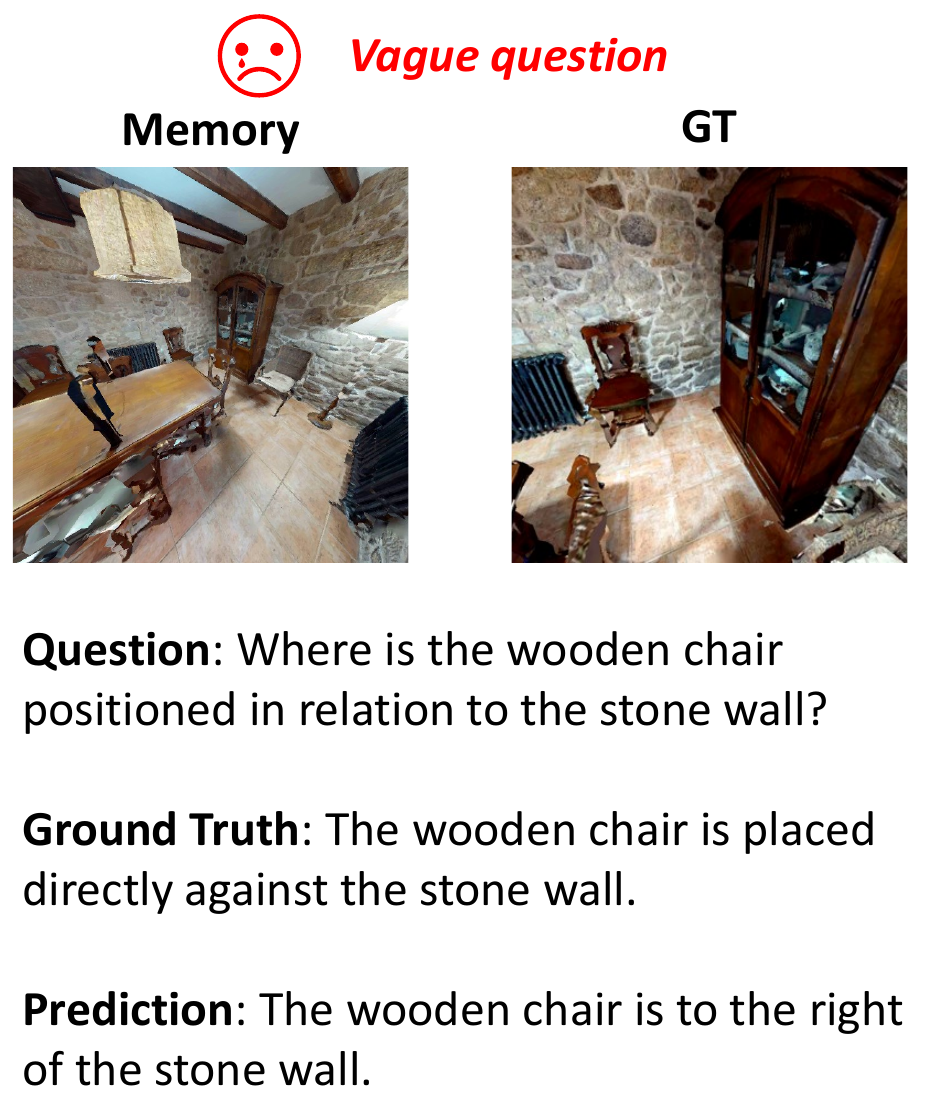}
% \vspace{-0.6em}
\caption{Failure cases.}
\label{fig:fail_1}
% \vspace{-1.2em}
\end{figure}

\begin{figure}[!t]
\centering
\includegraphics[width=0.8\linewidth]{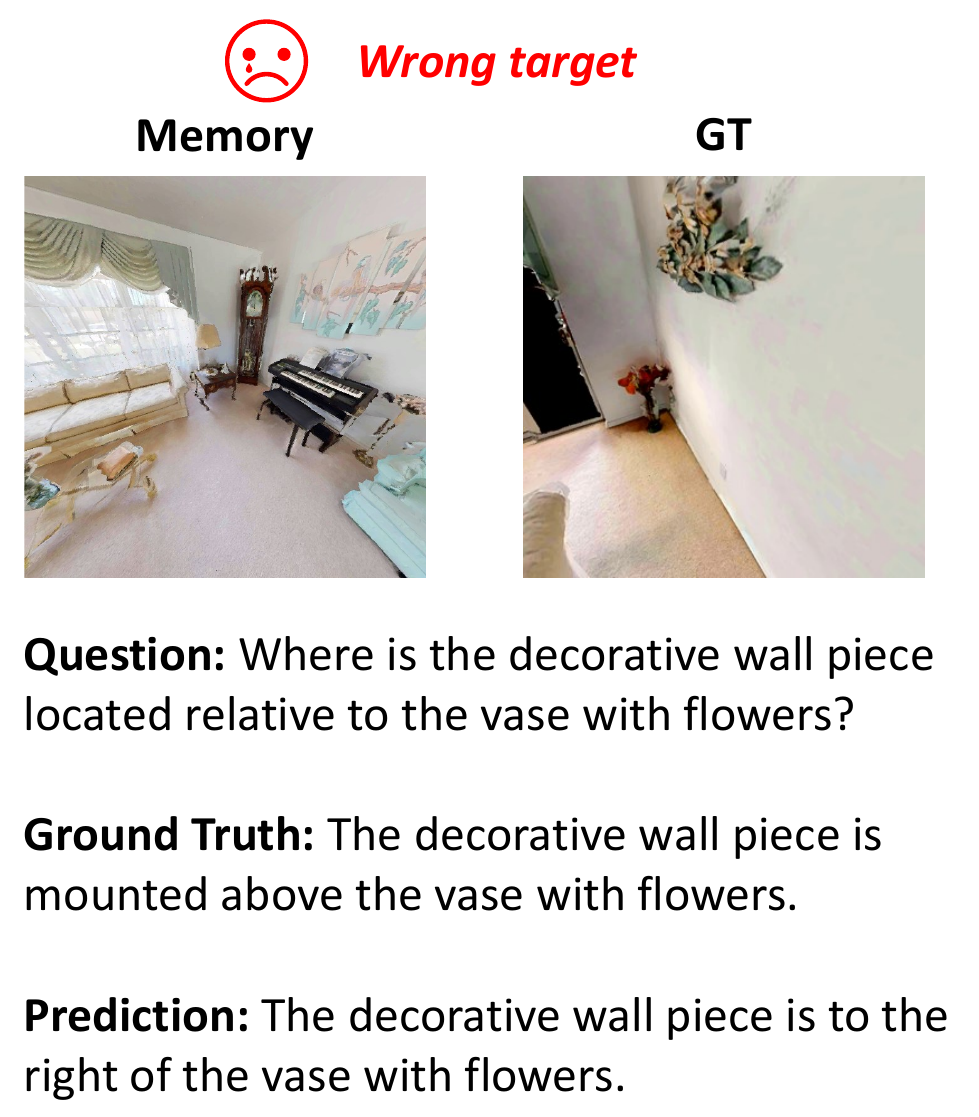}
% \vspace{-0.6em}
\caption{Failure cases.}
\label{fig:fail_2}
% \vspace{-1.2em}
\end{figure}

\begin{figure}[!t]
\centering
\includegraphics[width=0.8\linewidth]{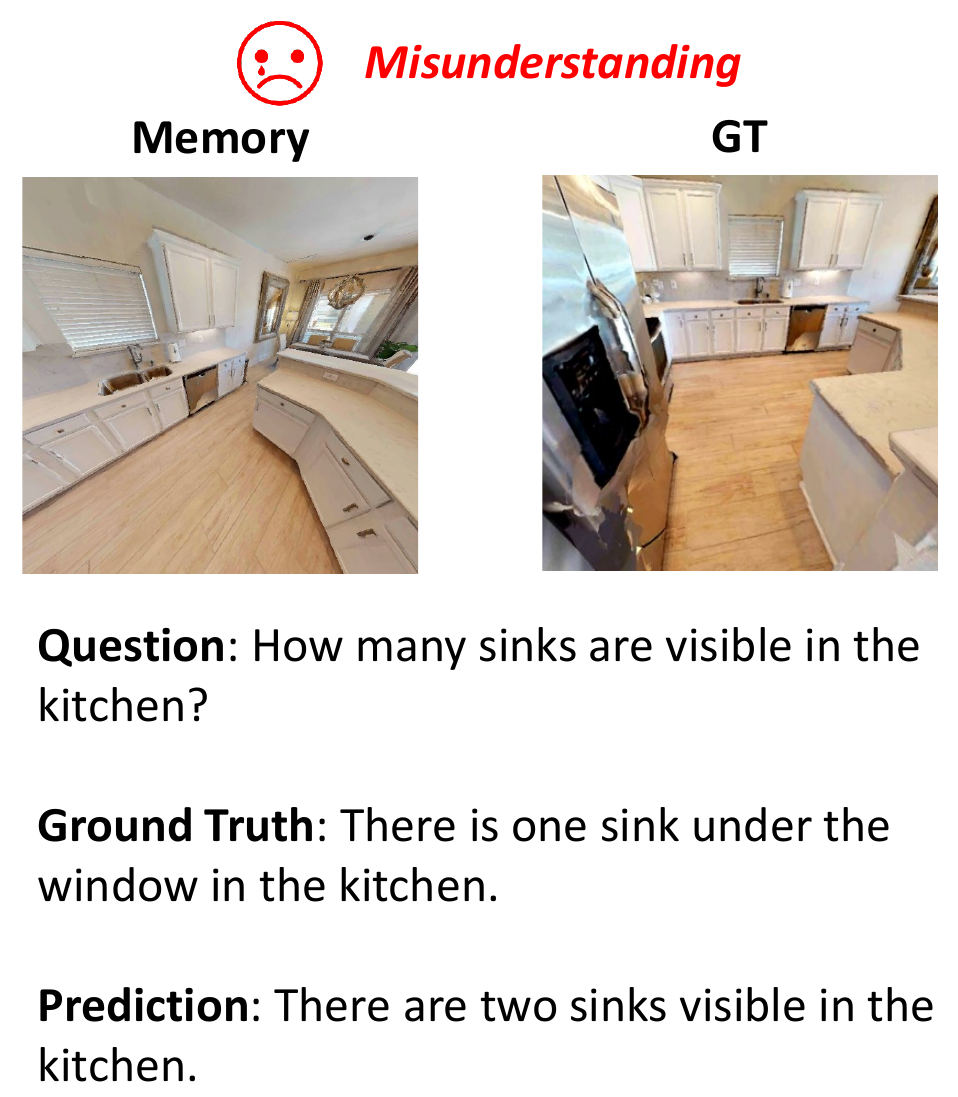}
% \vspace{-0.6em}
\caption{Failure cases.}
\label{fig:fail_3}
% \vspace{-1.2em}
\end{figure}

In addition to common navigation failures (such as selecting the wrong object in the memory or exceeding the maximum exploration steps), we further analyze the causes of memory-based question answering failures. First, ambiguities that inevitably arise during data generation, as in \cref{fig:fail_1}, may cause the model to retrieve the correct memory but still produce an incorrect answer. Second, due to the spatial understanding limitations of MLLMs, the agent may retrieve the wrong memory, as illustrated in \cref{fig:fail_2}, or generate an incorrect description even when the correct memory is retrieved, as shown in \cref{fig:fail_3}.

\section{Data Construction Details}
\begin{figure}[!t]
\centering
\includegraphics[width=1\linewidth]{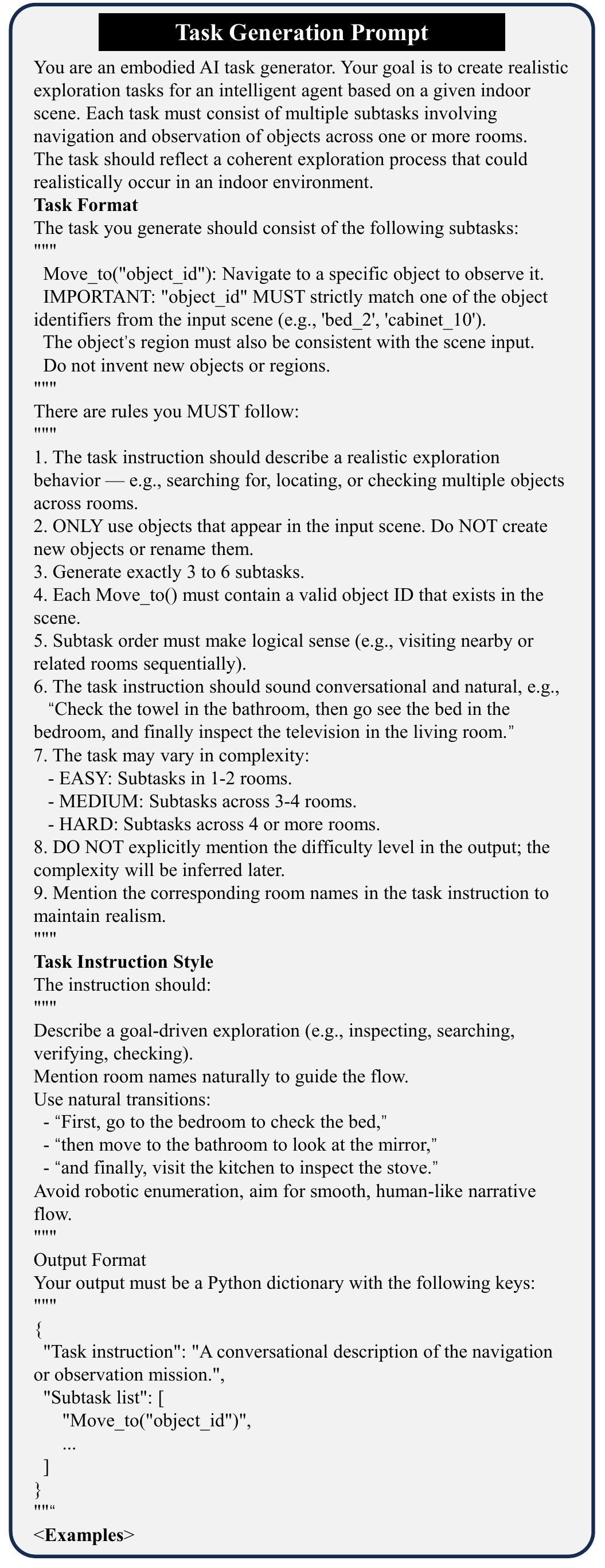}
\caption{Task instruction generation prompt.}
\label{fig:task}
\end{figure}

\begin{figure}[!t]
\centering
\includegraphics[width=1\linewidth]{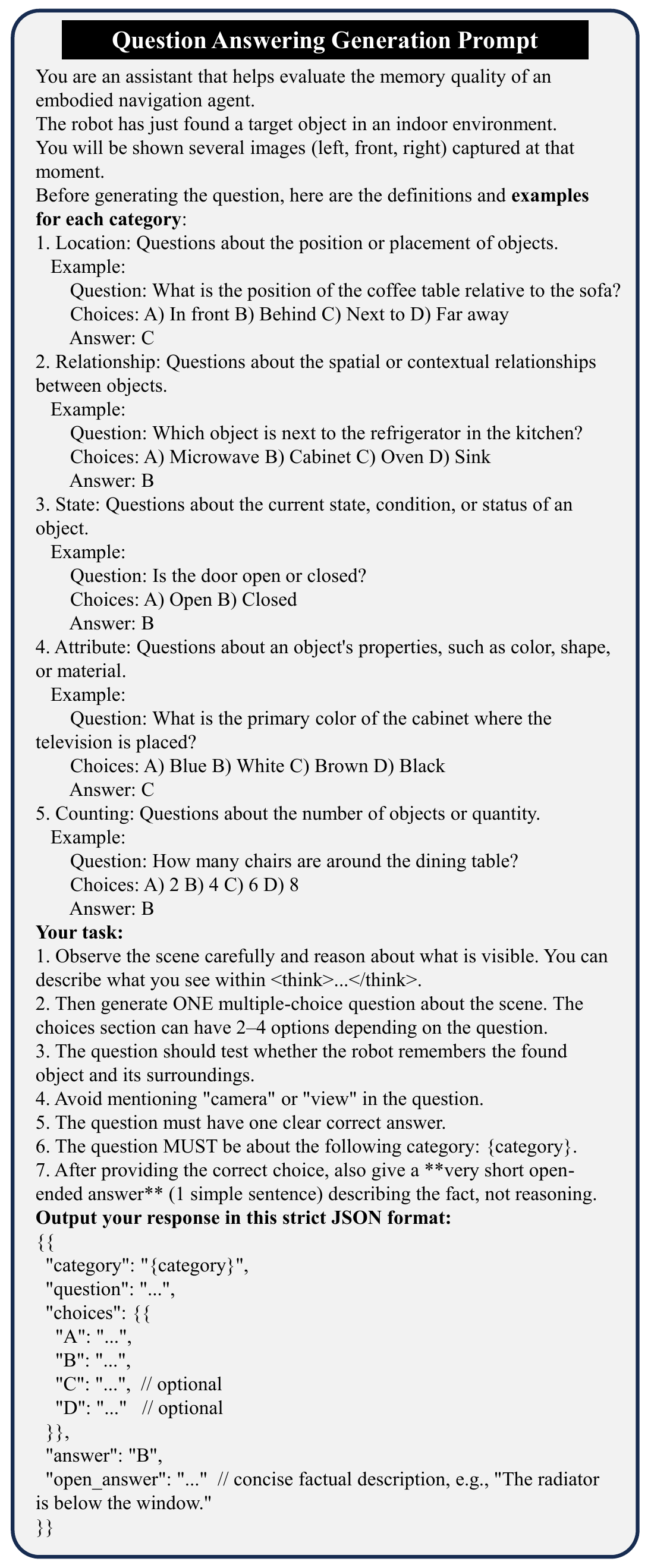}
\caption{Question answering generation prompt.}
\label{fig:qa}
\end{figure}

The HM3DSem includes labels for objects and their corresponding regions. We utilize the room names corresponding to each region provided in \cite{song2025towards}, such as bedroom, bathroom, etc. Finally, we input the object and its corresponding room information into LLM (Qwen3-235B-A22B-Instruct) and generate a multi-object navigation task based on the prompts in \cref{fig:task}. Since the illusion problem in LLM can generate non-existent or incorrect objects, we automatically filter out these erroneous task instructions when generating trajectories using Habitat-sim. Then, we input the observation images of successfully navigated targets in the trajectory into VLM (Qwen3-VL-235B-A22B-Instruct) to generate question-answer pairs, using prompts as shown in \cref{fig:qa}. \cref{tab:data_total} shows the specific statistical information of the trajectories and targets in our constructed dataset.

\noindent \textbf{Training Sample Construction.}
Since the trajectory data contains detailed information for each step, we designed a training sample construction \cref{alg:data_const} based on multimodal information. A task includes multiple trajectories corresponding to navigation trajectories for multiple target objects. The memory bank includes images, text, and location information. Since text generated based on an image tagging model is inaccurate, and location information is limited, similarity calculation mainly relies on image information. We set $\omega_o$, $\omega_f$, and $\omega_p$ to be 0.5, 0.3, and 0.2, respectively. However, memory updates based on similarity filtering found that they could not accurately collect goal-related observation images, \ie{}, correct memories. Therefore, we forcibly insert goal-related memories into each trajectory to ensure correct memory retrieval.

We use a 20-step action sampling interval to avoid high sample repetition, and a 10-step memory sampling interval to reduce the computational cost of memory retrieval during training. We calculate the mean and standard deviation of the similarity of the 10 most recent samples and dynamically filter context memories in each step based on an adaptive similarity threshold. The continuous action window is 6 steps to ensure action coherence. Finally, 11,684 samples are obtained as training data.

\section{Training Details}

\begin{figure*}[!t]
\centering
\includegraphics[width=1\linewidth]{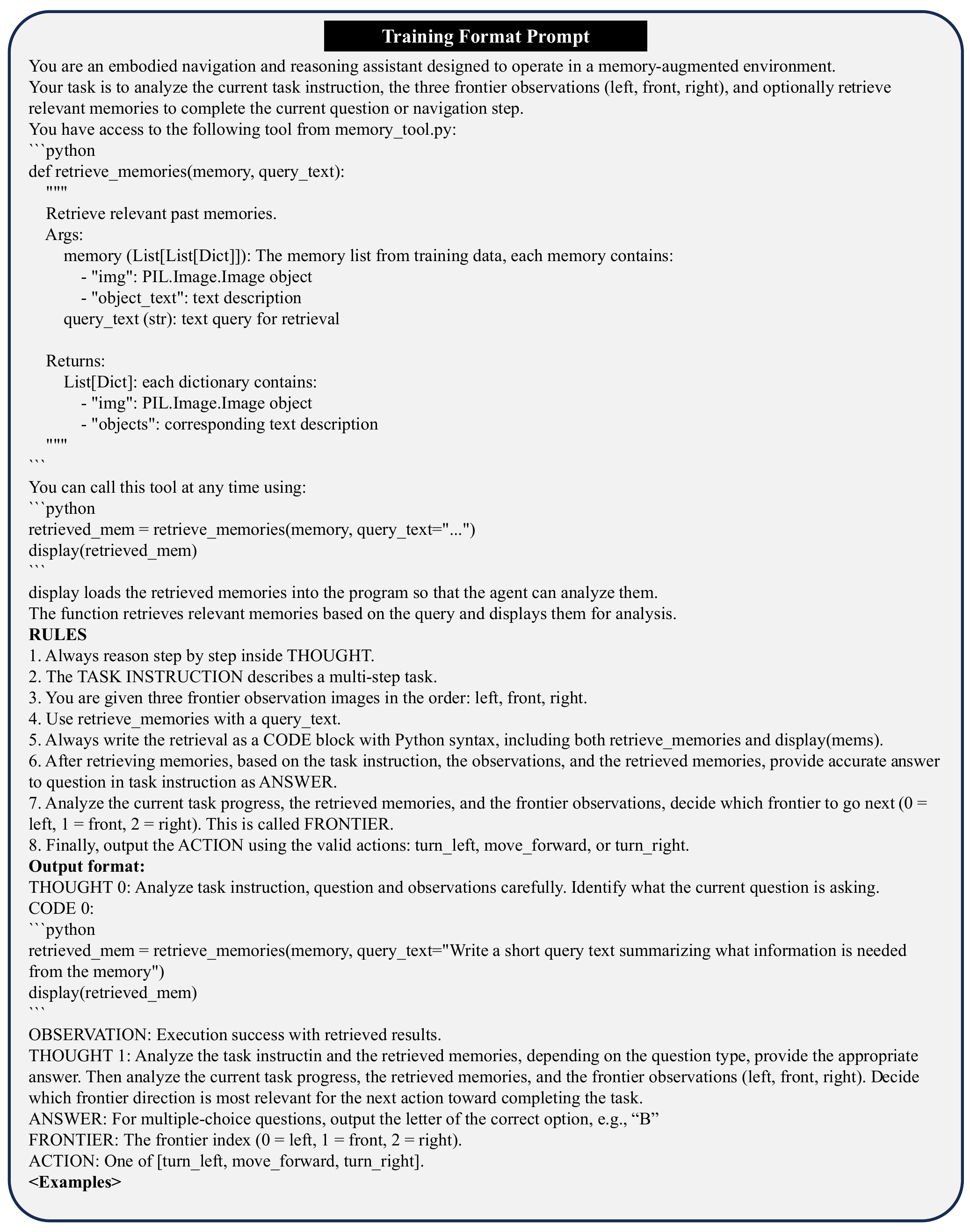}
% \vspace{-0.6em}
\caption{Training format prompt.}
\label{fig:train}
% \vspace{-1.2em}
\end{figure*}

We use EasyR1, a simplified version of the Verl framework. The specific training hyperparameters are shown in \cref{tab:train}. The training format prompt is shown in \cref{fig:train}. We use 8 NVIDIA H200 GPUs for training, which takes approximately 60 hours.

\section{Experimental Details}
\noindent \textbf{3D-Mem} is an embodied exploration method based on a multimodal large language model. It constructs a 3D memory bank by collecting multi-view observations. When performing embodied tasks, the agent builds front-end snapshots based on observations and depth images for exploration, and saves memory snapshots to find objects. Once the target object is confirmed in the memory snapshot, the agent stops running. The memory bank includes observed images and their corresponding object categories and masks. Due to the limitations of the MLLM context window, not all memory information can be input into MLLM simultaneously. It mitigates this problem by using object category-based relevance to filter memories.

\begin{figure}[!t]
\centering
\includegraphics[width=0.8\linewidth]{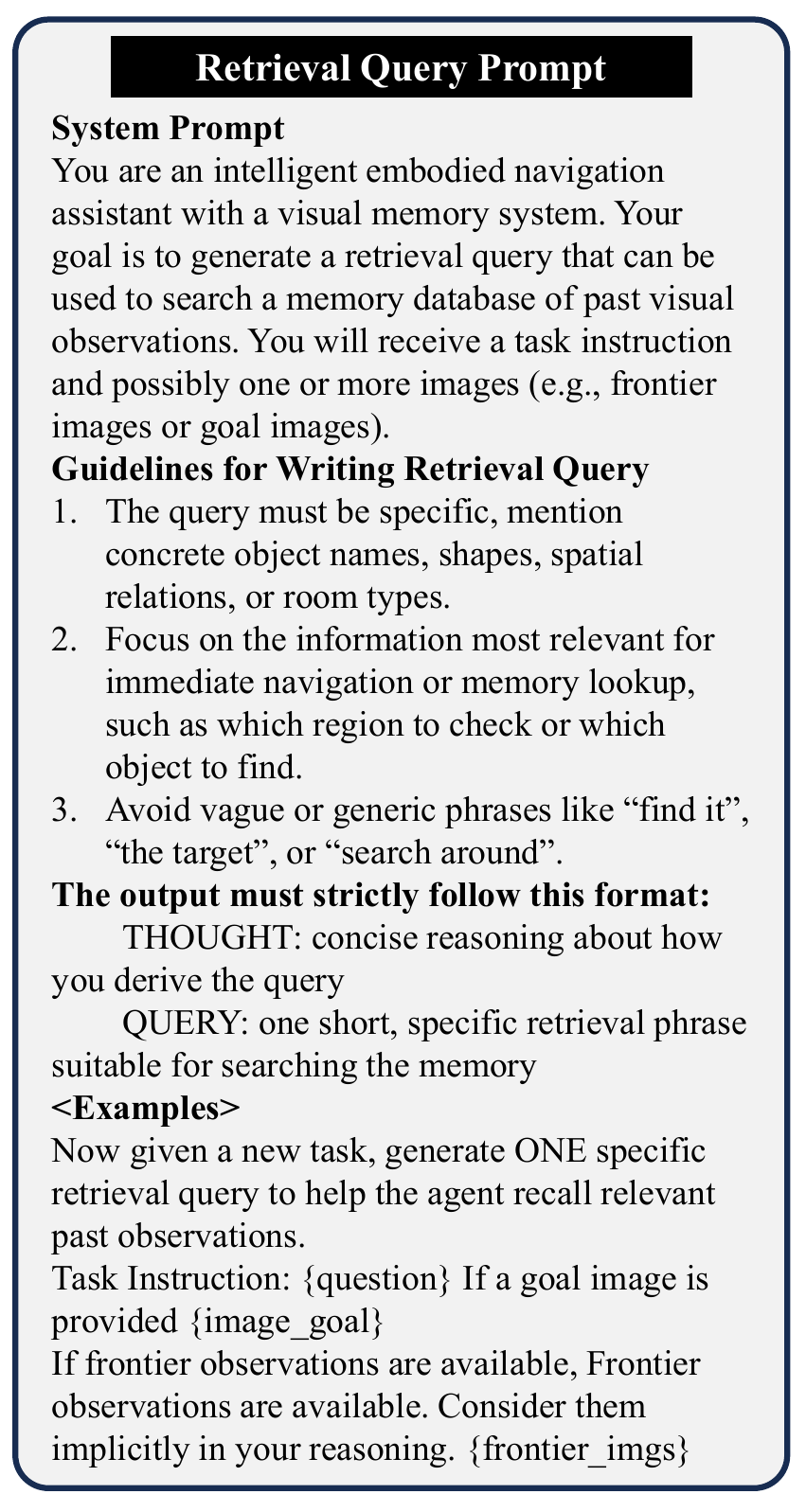}
% \vspace{-0.6em}
\caption{Query generation prompt.}
\label{fig:query}
% \vspace{-1.2em}
\end{figure}

\noindent \textbf{RA-Mem} is our embodied exploration method for actively retrieving memories, developed based on 3D-Mem~\cite{yang20253d}. The query prompt in \cref{fig:query} is input into MLLM to generate query text, and then retrieve the most relevant memories using feature similarity matching to help the model navigate and perform embodied question answering. This effectively improved model performance and reduced task completion time as shown in \cref{tab:full}.

\noindent \textbf{MemoryExplorer} builds upon the RA-Mem method, which utilizes only MLLM for inference, by introducing reinforcement learning fine-tuning. This allows for end-to-end training of an MLLM, enhancing its active memory retrieval and exploration capabilities. 

\begin{figure}[!t]
\centering
\includegraphics[width=1\linewidth]{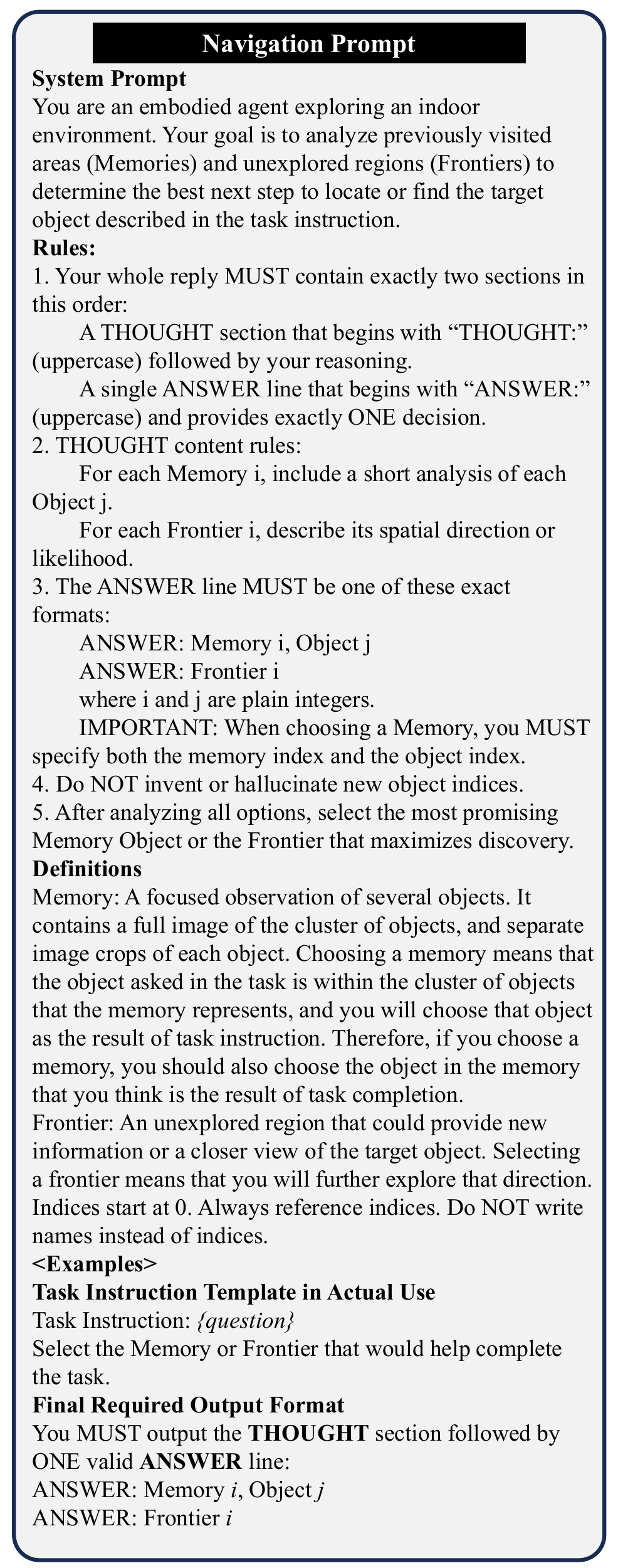}
% \vspace{-0.6em}
\caption{Navigation prompt.}
\label{fig:nav}
% \vspace{-1.2em}
\end{figure}

\begin{figure}[!t]
\centering
\includegraphics[width=1\linewidth]{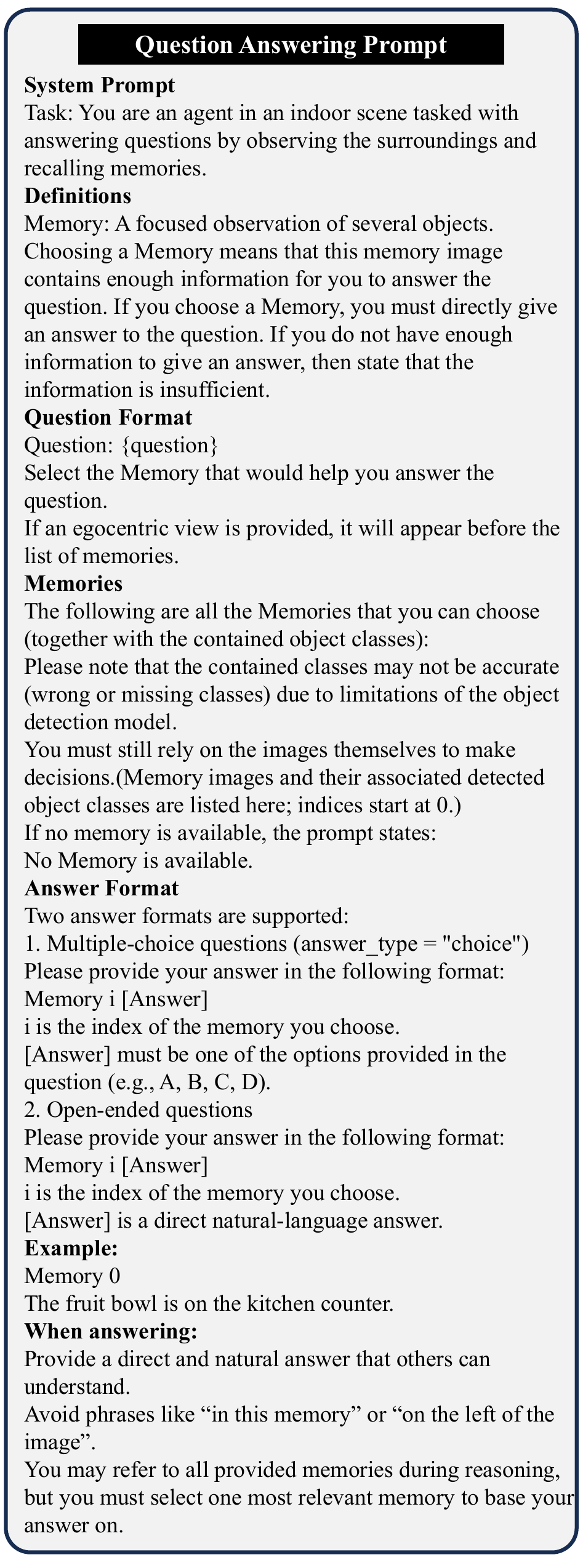}
% \vspace{-0.6em}
\caption{Memory-based question answering prompt.}
\label{fig:eqa}
% \vspace{-1.2em}
\end{figure}

\noindent \textbf{Embodied Exploration.} 
Similar to Goat-Bench~\cite{khanna2024goat}, an LMEE task consists of multiple subtasks. We define each subtask type as instance-level text description, image, and question-and-answer: ``The complete task is: \{task\_instruct\} Now you need to perform the subtask of finding the \{goal\_name\}, which is exactly described as the \{lang\_desc\}", ``The complete task is: \{task\_instruct\} Now you need to perform the subtask of finding the exact \{goal\_name\}, which is captured at the center of the following image? You need to pay attention to the environment and find the exact object.", and ``\{question\}". The agent first executes the multi-goal navigation task with navigation prompt as shown in \cref{fig:nav}, where each subtask includes the overall task instructions and task type description to drive the task progress. Then, the agent executes memory-based question-and-answer, using memory retrieval to answer the given questions with a question answering prompt as shown in \cref{fig:eqa}.

\noindent Most of our exploration settings follow 3D-Mem~\cite{yang20253d}, the frontier-based exploration framework is built upon the Explore-EQA~\cite{ren2024explore}, we maintain a 3D voxel occupancy map (0.1 m resolution) and update free space using depth observations and camera poses. The navigable region is defined as the free-voxel slice at 0.4 m height. Areas within 1.7 m of the agent’s trajectory are treated as explored; the rest remain unexplored.
Frontiers are formed by clustering pixels in the unexplored region using Density-Based Spatial Clustering of Applications with Noise (DBSCAN). A frontier $F = (r, p, I_{obs})$ includes the pixel cluster $r$, a navigable boundary point $p$, and an observation $I_{obs}$. We filter out small clusters ($<20$ pixels), update frontiers when the IoU with the previous version drops below 0.95, and split wide frontiers ($>150^\circ$ FOV) via K-means to improve navigation flexibility. Note that this voxel representation does not support multi-floor scenes.
For VLM prompting, only image observations are included. If the VLM chooses a frontier $F$, its associated location $p$ becomes the navigation target.
At each time step t, we collect 3 egocentric views with a 60° angular interval. The original views are captured at 1280 × 1280 resolution to improve object detection quality and then resized to 360 × 360 as input candidates for the VLM. Frontier snapshots are directly captured at 360 × 360.
We use YOLOv8x-World, implemented by Ultralytics, with a 200-class detection set from ScanNet. Each task is limited to a maximum of 50 steps. And a task completion condition where the agent's location is within 1m of the target object. We set the number of top\-k retrieved memories to 3, consistent with the training settings, while maintaining the MLLM's inference speed.

\section{Limitations}

The primary limitation of MLLM-based embodied exploration lies in its slow inference speed, which prevents real-time execution of embodied tasks. Developing more lightweight models will be an important direction for future research. In addition, the results on LMEE-Bench indicate that current methods are not yet able to effectively handle challenging embodied tasks that require long-term memory. Improving the accuracy and efficiency of long-term memory storage and retrieval will be crucial for advancing practical deployment.

\begin{table*}[!t]
\centering
\caption{Overall Statistics for trajectory and goal count across difficulty levels.}
\resizebox{0.9\linewidth}{!}{
    \begin{tabular}{lcccccccccc}
    \toprule
    & \multicolumn{3}{c}{\textbf{Trajectory Statistics}} 
    & \multicolumn{6}{c}{\textbf{Goal Statistics}} \\
    \cmidrule(lr){2-4} \cmidrule(lr){5-10}
    \textbf{Difficulty} 
    & \textbf{Tasks} & \textbf{Total Steps} & \textbf{Avg} 
    & \textbf{Train Total} & \textbf{Train Max} & \textbf{Train Min} & \textbf{Test Total}  & \textbf{Test Max} & \textbf{Test Min} & \textbf{Distance} \\
    \midrule
    All  & 1982 & 377311 & 190.37 & 8880 & 8 & 2  & 828 & 9 & 2 &1-30m  \\
    Easy  & 764  & 62961  & 82.41  & 2743 & 7 & 2 & 118  & 5 & 2 & 1-5m \\
    Medium & 1058 & 256598 & 242.53 & 5315 & 8 & 3 & 563  & 8 & 3 & 5-10m \\
    Hard  & 160 & 57752 & 360.95 & 822 & 8 & 5 & 147 & 9 & 5 & 10-30m \\
    \bottomrule
    \end{tabular}
}
\label{tab:data_total}
\end{table*}

\begin{algorithm}[!t]
\caption{Memory-Augmented Training Data Construction}
\label{alg:data_const}
\KwIn{Task set $\mathcal{D}$, CLIP encoder $\mathcal{E}$, sample interval $S$, memory interval $U$, continuous action window $W$}
\KwOut{Training dataset $\mathcal{T}$}
\ForEach{task $d \in \mathcal{D}$}{
    Load instruction, text, trials, and QA pairs\;
    Initialize memory bank $\mathcal{M}\leftarrow\varnothing$\;
    \ForEach{trial $k$ in order}{
        Load position $p$, get all images and text features $o$ and $f$ using $\mathcal{E}$\;
        Select a QA pair from past trials\;
        \For{$i=1$ to $T$}{
            \tcp{Dynamic memory update}
            \If{$i - i_{\text{last\_mem}} \ge U$}{
                Compute similarity between current memory $(p_c, f_c, o_c)$ and recent memory entries\;
                \If{Novelty condition satisfied}{
                    Append new memory entry to memory bank $\mathcal{M}$\;
                    Update $i_{\text{last\_mem}} \gets i$\;
                }
            }
            \tcp{Sample continuous action}
            \If{Continuous action window $W$ around $i$ contains too many distinct actions}{continue\;}
            \tcp{Sample training data}
            \If{$i-i_{\text{sample}}<S$}{continue\;}
            Build prompt with triplet images, instruction, memory hint, and QA\;
            Get next-action label $y_i$\;
            Append sample $(\text{prompt},\text{images},\mathcal{M},y_i, \text{answer})$ to $\mathcal{T}$\;
            Update $i_{\text{sample}}\leftarrow i$\;
        }
        \tcp{Goal-related memory}
        Add final-step memory entry to $\mathcal{B}$\;
    }
}
\Return $\mathcal{T}$\;
\end{algorithm}

\begin{table}[!t]
\centering
\caption{Hyperparameters Used in Training}
\resizebox{0.9\linewidth}{!}{
\begin{tabular}{lc}
\toprule
\textbf{Hyperparameter} & \textbf{Value} \\
\midrule
learning\_rate & 1e-6 \\
global\_batch\_size & 128 \\
rollout\_batch\_size & 256 \\
temperature & 1.0 \\
num\_generations & 5 \\
num\_train\_steps & 160 \\
max\_prompt\_length & 16384 \\
max\_response\_length & 4096 \\
batch\_size\_per\_device\_for\_update & 4 \\
batch\_size\_per\_device\_for\_experience & 8 \\
KL coefficient & 0.1 \\
Top-$k$ & 3 \\
consistency coefficient & 0.5 \\
Scaling factor & 1.2 for success / 0.5, 0.6 for fail \\
$w_{\text{act}}, w_{\text{front}}, w_{\text{ans}}, w_{\text{fmt}}$ 
& $0.2,\, 0.2,\, 0.4,\, 0.2$ \\
\bottomrule
\end{tabular}
}
\label{tab:train}
\end{table}

\end{document}